\documentclass[10pt,twocolumn,letterpaper]{article}

\usepackage{iccv}
\usepackage{times}
\usepackage{epsfig}
\usepackage{graphicx}
\usepackage{amsmath}
\usepackage{amssymb}

\usepackage{array}
\usepackage{booktabs}
\usepackage{subcaption}
\usepackage{pifont}
\usepackage[font=small]{caption}
\usepackage{url}
\usepackage{xspace} 
\usepackage{xcolor}
\usepackage{CJKutf8}
\usepackage{multirow}
\usepackage[utf8]{inputenc}

\newcommand\samethanks[1][\value{footnote}]{\footnotemark[#1]}

\newcommand{\vatex}{\textsc{VaTeX}\xspace}
\newcommand{\vatexen}{\textsc{VaTeX}-en\xspace}
\newcommand{\vatexzh}{\textsc{VaTeX}-zh\xspace}

\definecolor{darkgreen}{RGB}{84,174,50}
\newcommand{\up}[1]{$\textcolor{darkgreen}{(+#1)}$}

\newcommand{\en}{$\textcolor{white}{(+0.00)}$}

\newcommand{\confnum}[2]{#1 $\pm${\scriptsize #2}}

\usepackage[breaklinks=true,bookmarks=false,colorlinks]{hyperref}

\iccvfinalcopy 

\def\iccvPaperID{****} 

\ificcvfinal\fi
\begin{document}

\title{\vatex: A Large-Scale, High-Quality Multilingual Dataset \\
for Video-and-Language Research \\
\href{http:\\vatex-challenge.org}{vatex-challenge.org}
}

\author{Xin Wang\thanks{Equal contribution.}~$^1$ \quad Jiawei Wu\samethanks~$^1$ \quad Junkun Chen$^2$ \quad Lei Li$^2$ \quad Yuan-Fang Wang$^1$ \quad William Yang Wang$^1$\\
$^1$University of California, Santa Barbara, CA, USA\\
$^2$ByteDance AI Lab, Beijing, China\\
}

\maketitle

\begin{abstract}
We present a new large-scale multilingual video description dataset, \vatex\footnote{\vatex~stands for Video And TEXt, where X also represents various languages.}, which contains over $41,250$ videos and $825,000$ captions in both English and Chinese. Among the captions, there are over $206,000$ English-Chinese parallel translation pairs. Compared to the widely-used MSR-VTT dataset~\cite{msrvtt}, \vatex is multilingual, larger, linguistically complex, and more diverse in terms of both video and natural language descriptions.
We also introduce two tasks for video-and-language research based on \vatex: (1) Multilingual Video Captioning, aimed at describing a video in various languages with a compact unified captioning model, 
and (2) Video-guided Machine Translation, to translate a source language description into the target language using the video information as additional spatiotemporal context. 
Extensive experiments on the \vatex dataset show that, first, the unified multilingual model can not only produce both English and Chinese descriptions for a video more efficiently, but also offer improved performance over the monolingual models. Furthermore, we demonstrate that the spatiotemporal video context can be effectively utilized to align source and target languages and thus assist machine translation.
In the end, we discuss the potentials of using \vatex for other video-and-language research.
\end{abstract}

\begin{figure}
    \centering
    \begin{subfigure}{.48\textwidth}
        \includegraphics[width=\linewidth]{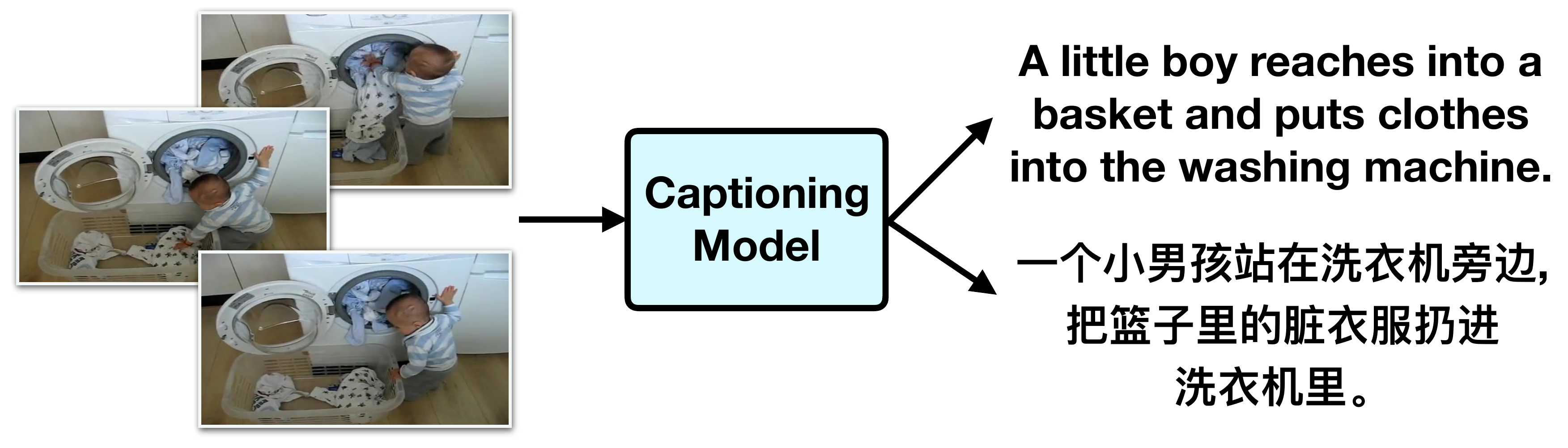}
        \caption{Multilingual Video Captioning}
        \label{fig:demo-cap}
    \end{subfigure} \\
    \begin{subfigure}{.5\textwidth}
        \includegraphics[width=\linewidth]{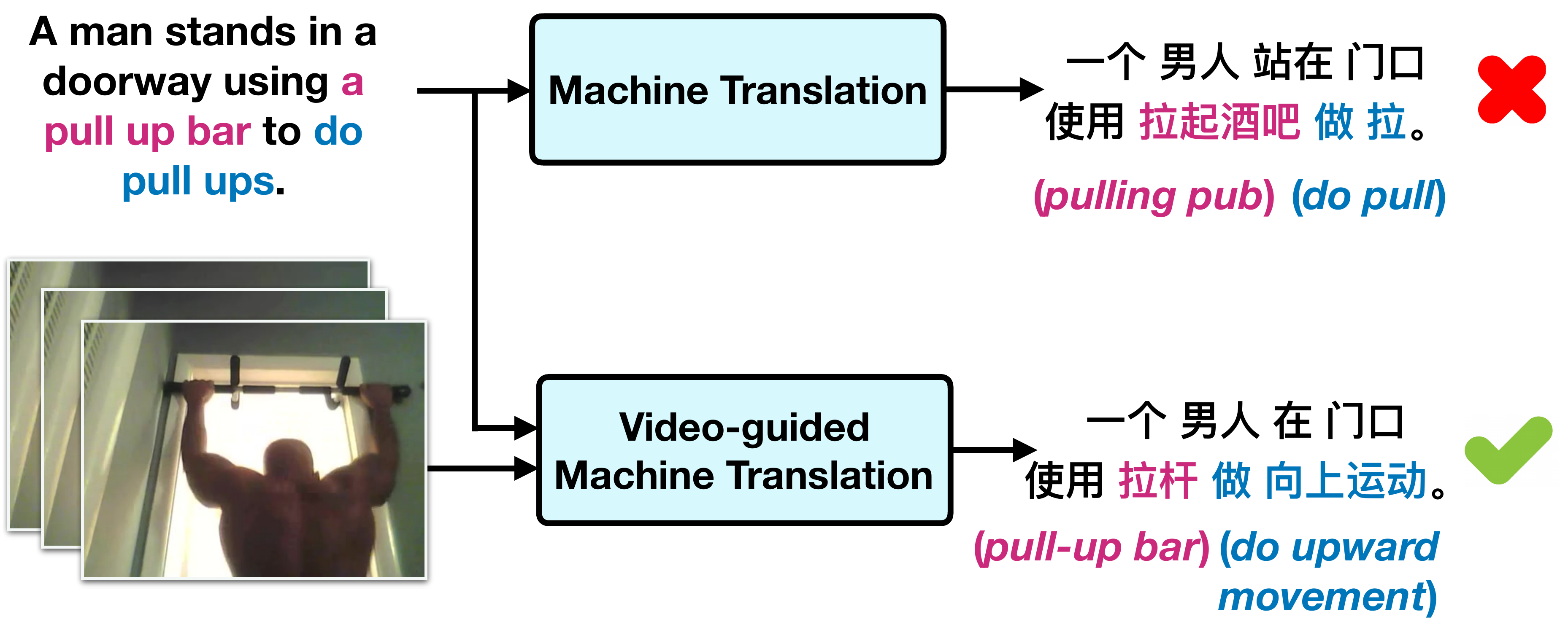}
        \caption{Video-guided Machine Translation}
        \label{fig:demo-mmt}
    \end{subfigure}
    \caption{Demonstration of the \vatex tasks. (a) A compact unified video captioning model is required to accurately describe the video content in both English and Chinese. (b) The machine translation model mistakenly interprets ``\textit{pull up bar}" as ``\textit{pulling pub}" and ``\textit{do pull ups}" as ``\textit{do pull}" (two verbs), which are meaningless. While with the relevant video context, the English sentence is precisely translated into Chinese.}
\end{figure}

\section{Introduction}
Recently, researchers in both computer vision and natural language processing communities are striving to bridge videos and natural language. For a deeper understanding of the activities, the task of video captioning/description aims at describing the video content with natural language. A few datasets have been introduced for this task and cover a variety of domains, such as cooking~\cite{youcook,youcook2}, movie~\cite{rohrbach2015dataset}, human actions~\cite{msvd,msrvtt}, and social media~\cite{videostory}. Despite the variants of this task, the fundamental challenge is to accurately depict the important activities in a video clip, which requires high-quality, diverse captions that describe a wide variety of videos at scale.
Moreover, existing large-scale video captioning datasets are mostly monolingual (English only) and thus the development of video captioning models is restricted to English corpora. However, the study of multilingual video captioning is essential for a large population on the planet who cannot speak English.

\begin{figure*}
    \centering
    \includegraphics[width=\textwidth]{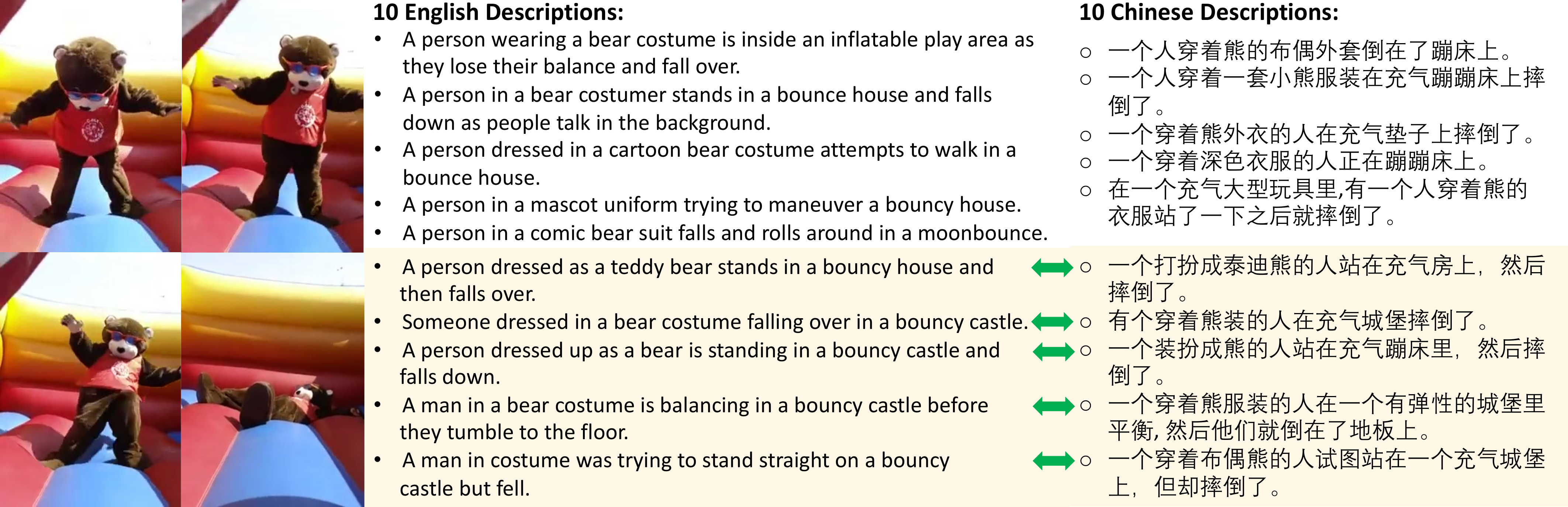}
    \caption{A sample of our \vatex dataset. The video has 10 English and 10 Chinese descriptions. All depicts the same video and thus are distantly parallel to each other, while the last five are the paired translations to each other.}
    \label{fig:sample}
\end{figure*}

To this end, we collect a new large-scale multilingual dataset for video-and-language research, \vatex, that contains over $41,250$ unique videos and $825,000$ high-quality captions. It covers $600$ human activities and a variety of video content. Each video is paired with $10$ English and $10$ Chinese diverse captions from $20$ individual human annotators. Figure~\ref{fig:sample} illustrates a sample of our \vatex dataset.
Compared to the most popular large-scale video description dataset MSR-VTT~\cite{msrvtt}, \vatex is characterized by the following major unique properties. First, it contains both English and Chinese descriptions at scale, which can support many multilingual studies that are constrained by monolingual datasets. Secondly, \vatex has the largest number of clip-sentence pairs with each video clip annotated with multiple unique sentences, and every caption is unique in the whole corpus. Thirdly, \vatex contains more comprehensive yet representative video content, covering $600$ human activities in total. Furthermore, both the English and Chinese corpora in \vatex are lexically richer and thus can empower more natural and diverse caption generation.

With the capabilities of the \vatex dataset, we introduce the task of multilingual video captioning (see Figure~\ref{fig:demo-cap}), which is to train a unified model to generate video descriptions in multiple languages (\eg, English and Chinese). 
However, \textit{would the multilingual knowledge further reinforce video understanding?} 
We examine different multilingual models where different portions of the architectures are shared for multiple languages. 
Experiments show that a compact unified multilingual captioning model is not only more efficient but also more effective than monolingual models. 

Video captioning is designed to push forward video understanding with natural language descriptions, but \textit{can video information help natural language tasks like machine translation in return}? To answer this question, we collect around 206K English-Chinese parallel sentences among all the captions and introduce a new task, video-guided machine translation (VMT), to translate a source language description into the target language using the video information as additional spatiotemporal context. We assume that the spatiotemporal context would reduce the ambiguity of languages (especially for verbs and nouns) and hence promote the alignment between language pairs. So we further conduct extensive experiments and verify the effectiveness of VMT. In Figure~\ref{fig:demo-mmt}, we demonstrate an example where video information can play a crucial role in translating essential information. 

In summary, our contributions are mainly three-fold:
\begin{itemize}
    \item We collect a new large-scale and high-quality multilingual video description dataset for the advance of the video-and-language research, and conduct in-depth comparisons among MSR-VTT, \vatex English corpus, and \vatex Chinese corpus.
    \item We introduce the task of multilingual video captioning and validate its efficiency and effectiveness of generating video descriptions in both English and Chinese with a compact, unified model. 
    \item We are the first to propose the task of video-guided machine translation and examine the effectiveness of incorporating spatiotemporal context to improve the performance of machine translation.
\end{itemize}

\section{Related Work}
\noindent\textbf{Video Description Datasets.}
Various datasets for video description/captioning have been introduced to empower different ways to describe the video content, covering a wide range of domains, such as cooking~\cite{youcook,youcook2,regneri2013grounding,tacos-mlevel}, movie~\cite{M-VAD,rohrbach2015dataset,Rohrbach2017}, social media~\cite{videostory}, and human activities~\cite{msvd,charades,msrvtt,krishna2017dense}. In Table~\ref{tab:overall}, we summarize existing video description datasets~\cite{Aafaq2018VideoDA} and briefly compare their major statistics. Generally, video description tasks can mainly be divided into two families, single-sentence generation (\eg, \cite{msvd,msrvtt}) and multi-sentence generation (\eg, \cite{krishna2017dense}), though they may appear as different variants due to the difference of the corpora, \eg, video title generation~\cite{VTW} and video story generation~\cite{videostory}.
In this work, we present a large-scale, high-quality multilingual benchmark for single-sentence generation, aiming at encouraging fundamental approaches towards a more in-depth understanding of human actions. As shown in Table~\ref{tab:overall}, our \vatex dataset is the largest benchmark in terms of video coverage and the language corpora; it also provides $20$ captions for each video clip to take into consideration human variance when describing the same video and hence supports more human-consistent evaluations. 
Moreover, our \vatex dataset contains both English and Chinese descriptions at scale, which is an order of magnitude larger than MSVD~\cite{msvd}. Besides, MSVD does not have any translation pairs as \vatex does. Therefore, \vatex can empower many multilingual, multimodal research that requires large-scale training.

\begin{table}
\setlength{\tabcolsep}{2pt}
    \centering
    \resizebox{0.5\textwidth}{!}{
        \begin{tabular}{l c c c c c c}
        \toprule
             \textbf{Dataset} & \textbf{MLingual} & \textbf{Domain} & \textbf{\#classes} & \textbf{\#videos:clips} & \textbf{\#sent} & \textbf{\#sent/clip}\\
             \midrule
             TACoS\cite{regneri2013grounding} & - & cooking & 26 & 127:3.5k & 11.8k & -\\
             TACoS-MLevel\cite{tacos-mlevel} & - & cooking & 67 & 185:25k & 75k & 3\\
             Youcook\cite{youcook} & - & cooking & 6 & 88:- & 2.7k & - \\
             Youcook II\cite{youcook2} & - & cooking & 89 & 2k:15.4k & 15.4k & 1\\
             MPII MD\cite{rohrbach2015dataset} & - & movie & - &  94:68k &  68.3k & 1 \\
             M-VAD\cite{M-VAD} & - & movie & - &  92:46k & 55.9k & - \\
             LSMDC\cite{Rohrbach2017} & - & movie & - &  200:128k & 128k & 1\\
             Charades\cite{charades} & - & indoor & 157 & 10k:10k & 27.8k & 2-3\\
             VideoStory\cite{videostory} & - & social media & - & 20k:123k & 123k & 1 \\
             ActyNet-Cap\cite{krishna2017dense} & - & open & 200 & 20k:100k & 100k & 1\\
             MSVD\cite{msvd} & \checkmark & open & - & 2k:2k & 70k & 35\\
             TGIF\cite{tgif} & - & open & - & -:100k & 128k & 1\\
             VTW\cite{VTW} & - & open & - & 18k:18k & 18k & 1 \\
             MSR-VTT\cite{msrvtt} & - & open & 257 & 7k:10k & 200k & 20\\
             \midrule
             \vatex (ours) & \checkmark & open & 600 & 41.3k:41.3k & 826k & 20 \\
        \bottomrule
        \end{tabular}
    }
    \caption{Comparison of the video description datasets.}
    \label{tab:overall}
\end{table}

\noindent\textbf{Multilingual Visual Understanding.}
Numerous tasks have been proposed to combine vision and language to enhance the understanding of either or both, such as video/image captioning~\cite{Donahue2015LongTermRC,Vinyals2015ShowAT,nocaps}, visual question answering (VQA)~\cite{Antol2015VQAVQ}, and natural language moment retrieval~\cite{Hendricks2017LocalizingMI}, etc. Multilingual studies are rarely explored in the vision and language domain. Gao \etal~\cite{Gao2015AreYT} introduce a multilingual image question answering dataset, and Shimizu \etal~\cite{Shimizu2018VisualQA-ja} propose a cross-lingual method for making use of English annotations to improve a Japanese VQA system. Pappas \etal~\cite{Pappas2016MultilingualVS} propose multilingual visual concept clustering to study the commonalities and differences among different languages. Meanwhile, multilingual image captioning is introduced to describe the content of an image with multiple languages~\cite{Lan2017FluencyGuidedCI,Tsutsui2017UsingAT,Li2018COCOCNFC}.  
But none of them study the interaction between videos and multilingual knowledge.  
Sanabria \etal~\cite{how2} collect English$\rightarrow$Portuguese subtitles for the automatic speech recognition (ASR) task, which however do not directly describe the video content.
Therefore, we introduce the \vatex dataset and the task of multilingual video captioning to facilitate multilingual understanding of video dynamics. 

\noindent\textbf{Multimodal Machine Translation.}
The multimodal machine translation task aims at generating a better target sentence by supplementing the source sentence with extra information gleaned from other modalities. Previous studies mainly focus on using images as the visual modality to help machine translation~\cite{specia2016shared,elliott-EtAl:2017:WMT,barrault2018findings}. The Multi30K dataset~\cite{elliott2016multi30k}, which is annotated based on the image captioning dataset Flickr30K~\cite{plummer2015flickr30k}, is commonly used in this direction. For instance, \cite{huang2016attention,gronroos2018memad} consider the object features of the images, and ~\cite{caglayan2016multimodal,libovicky2017attention} import convolutional image features into machine translation. Additionally, other studies~\cite{ma2017osu,calixto2017incorporating,madhyastha2017sheffield,caglayan2017lium} explore the cross-modal feature fusion of images and sentences.
In this work, we are the first to consider videos as the spatiotemporal context for machine translation and introduce a new task---video-guided machine translation.
Compared with images, videos provide richer visual information like actions and temporal transitions, which can better assist models in understanding and aligning the words/phrases between the source and target languages.
Moreover, the parallel captions in \vatex go beyond of spatial relations and are more linguistically complex than Multi30K, \eg, a series of actions.
Last but not least, our \vatex dataset contains over $206$K English-Chinese sentence pairs ($5$ per video), which is approximately seven times larger than Multi30K.

\section{\vatex Dataset}

\begin{table}
\small
\setlength{\tabcolsep}{4pt}
    \centering
    \begin{tabular}{l | c c c c}
    \toprule
        Split    & train & validation & public test & secret test \\
        \midrule
        \#videos & 25,991 & 3,000 & 6,000 & 6,278\\
        \#captions & 519,820 & 60,000 & 120,000 & 125,560\\
        action label & \checkmark & \checkmark & - & - \\
    \bottomrule
    \end{tabular}
    \caption{The splits of the \vatex dataset (\checkmark~indicates the videos have publicly accessible action labels). For the secret test set, we holdout the human-annotated captions for challenge use.}
    \label{table:split}
\end{table}

\begin{figure*}
    \centering
    \begin{subfigure}{0.33\textwidth}
    \includegraphics[width=\textwidth,height=3.8cm]{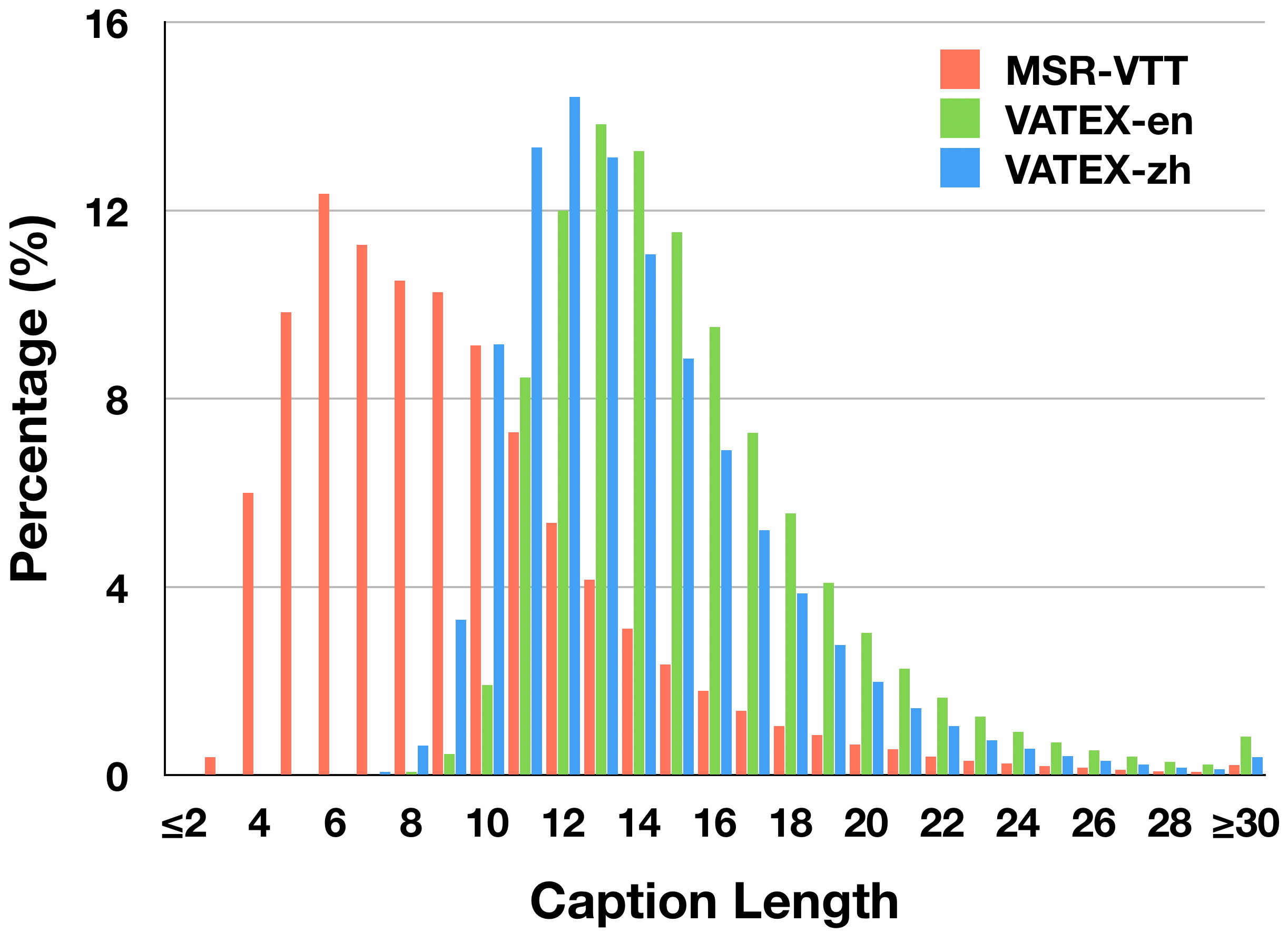}
    \caption{Distributions of caption lengths.}
    \label{fig:cap-length}
    \end{subfigure}
    \begin{subfigure}{0.33\textwidth}
        \includegraphics[width=\textwidth, height=3.8cm]{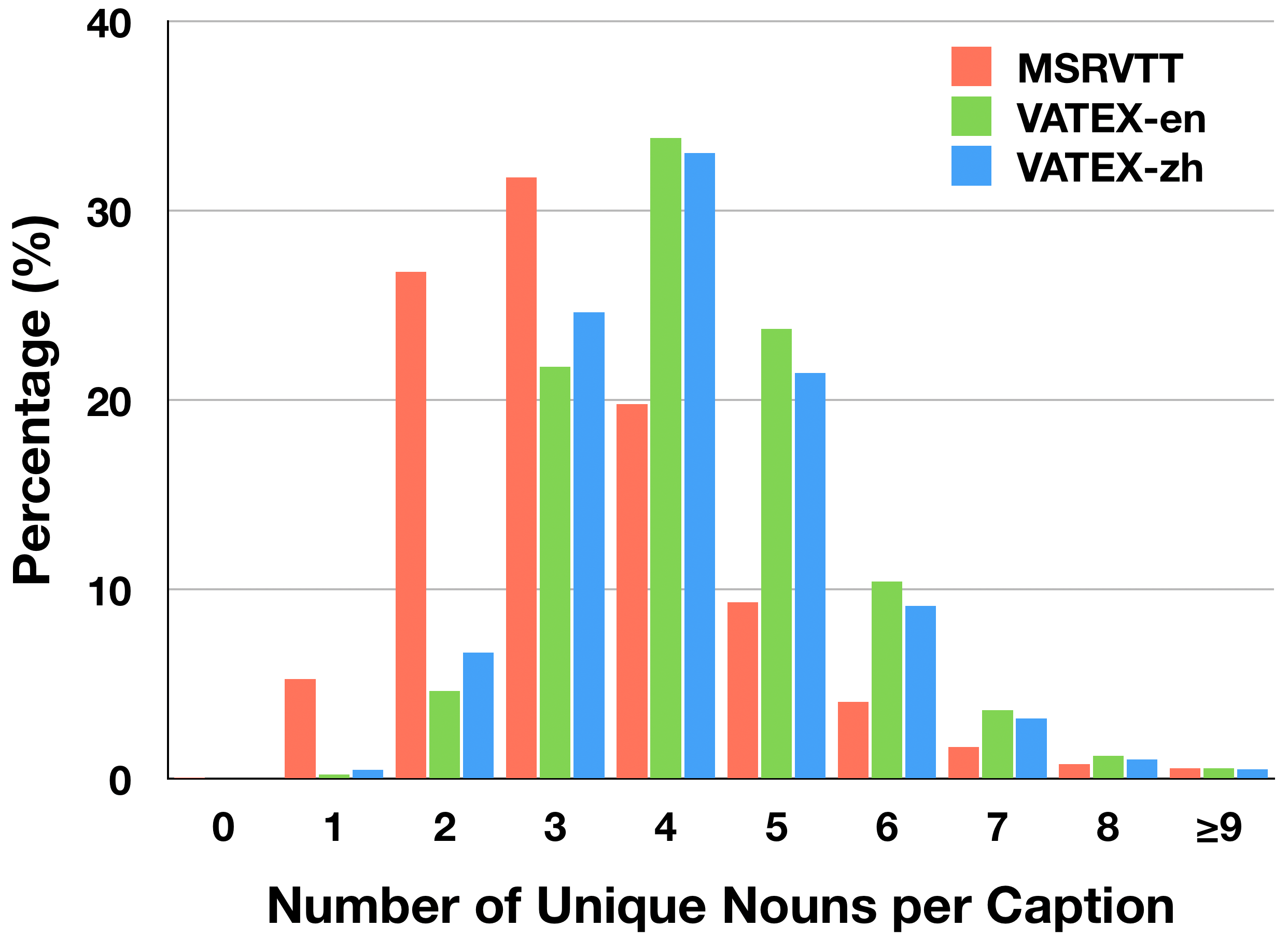}
        \caption{Distributions of unique nouns per caption.}
        \label{fig:noun-distribution}
    \end{subfigure}
    \begin{subfigure}{0.33\textwidth}
        \includegraphics[width=\textwidth, height=3.8cm]{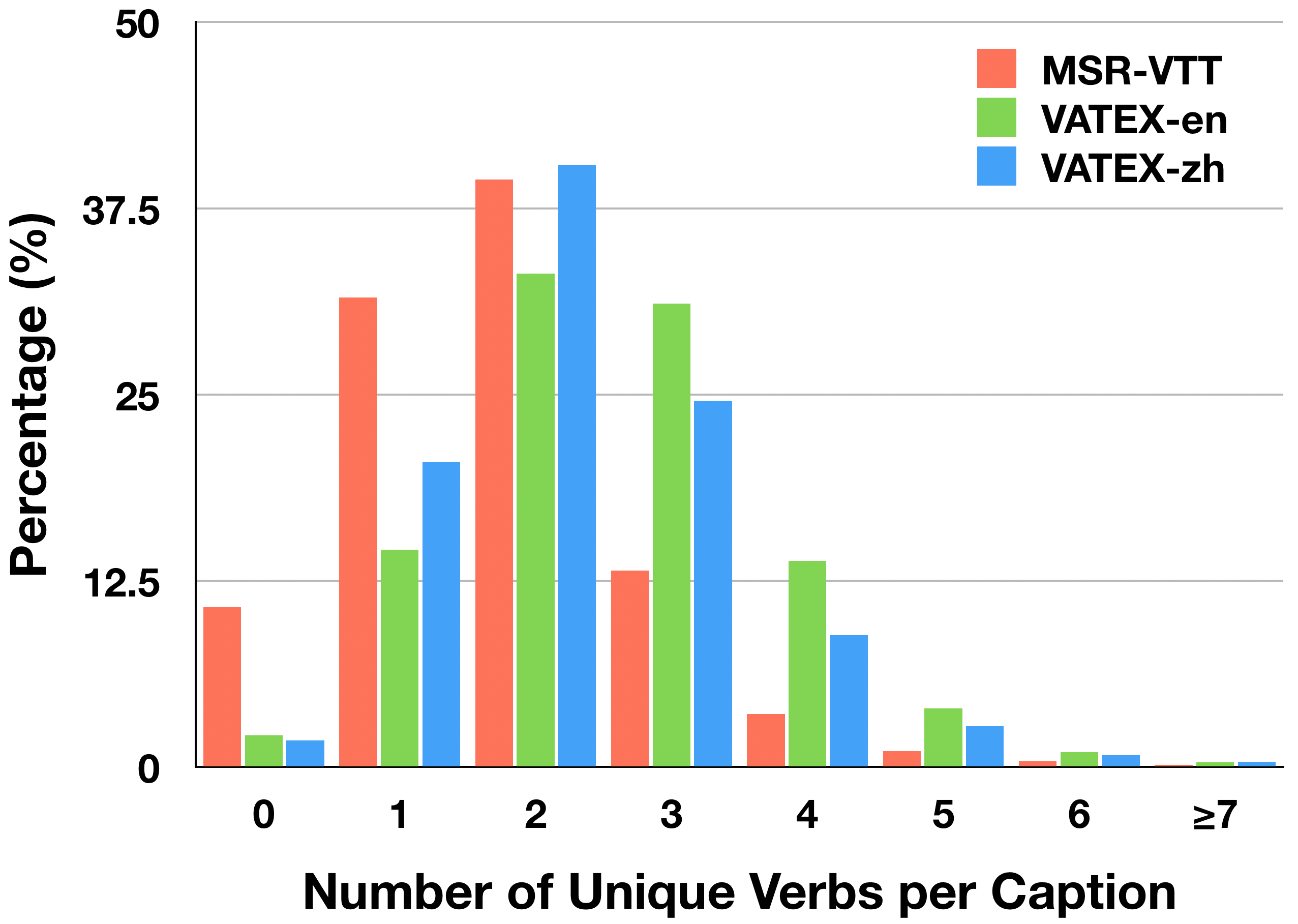}
        \caption{Distributions of unique verbs per caption.}
        \label{fig:verb-distribution}
    \end{subfigure}
    \caption{Statistical histogram distributions on MSR-VTT, \vatexen, and \vatexzh. Compared to MSR-VTT, the \vatex dataset contains longer captions, each with more unique nouns and verbs.}
    \label{fig:cap-per-cap}
\end{figure*}

\subsection{Data Collection}
For a wide coverage of human activities, we reuse a subset of the videos from the Kinetics-$600$ dataset~\cite{kay2017kinetics}, the largest and widely-used benchmark for action classification. Kinetics-$600$ contains $600$ human action classes and around half a million video clips. To collect those videos, Kay \etal~\cite{kay2017kinetics} first built an action list by combining previous video datasets~\cite{Heilbron2015ActivityNetAL,Kuehne2011HMDBAL,Soomro2012UCF101AD,Andriluka20142DHP,Wang2016ActionsT}, and then searched the videos from YouTube for candidates, which eventually were filtered by Amazon Mechanical Turkers. 
Each clip lasts around $10$ seconds and is taken from a unique YouTube video. 
The \vatex dataset connects videos to natural language descriptions rather than coarse action labels. Notably, we collect the English and Chinese descriptions of $41,269$ valid video clips from the Kinetics-$600$ validation and holdout test sets, costing approximately \$$51,000$ in total. The data collection window is around two months.
We have obtained approvals from the institutional reviewing agency to conduct human subject crowdsourcing experiments, and our payment rate is reasonably high (the estimated hourly rate is higher than the minimum wage required by law).

We split those videos into four sets as shown in Table~\ref{table:split}. Note that the train and validation sets are split from the Kinetics-$600$ validation set, and the test sets are from the Kinetics-$600$ holdout test set. Below we detail the collection process of both English and Chinese descriptions.

\subsubsection{English Description Collection}
Towards large-scale and diverse human-annotated video descriptions, we build upon Amazon Mechanical Turk (AMT)\footnote{\url{https://www.mturk.com}} and collect $10$ English captions for every video clip in \vatex, where each caption from an individual worker. 
Specifically, the workers are required to watch the video clips and describe the corresponding captions in English. 
In each assignment, the workers are required to describe $5$ videos.
We show the instructions that the workers should describe all the important people and actions in the video clips with the word count in each caption no less than $10$. The AMT interface can be found in the supplementary material, which contains more details.

To ensure the quality of the collected captions, we employ only workers from the English-speaking countries, including Australia, Canada, Ireland, New Zealand, UK, and USA. 
The workers are also required to complete a minimum of $1$K previous tasks on AMT with at least a $95\%$ approval rate.
Furthermore, we daily spot-check the captions written by each worker to see if they are relevant to the corresponding videos. Meanwhile, we run scripts to check the captions according to the following rules: (1) whether the captions are shorter than $8$ words; (2) whether there are repeated captions; (3) whether the captions contain sensitive words; and (4) whether the captions are not written in English. We reject all the captions that do not achieve the requirements and block the workers consistently providing low-quality annotations. The rejected captions are re-collected until all captions strictly follow the requirements.
In preliminary experiments, we find that the workers may struggle to write good captions with only the instructions. Hence, we further provide some accepted good examples and rejected bad examples (both are unrelated to the current video clips) for workers' reference. We observe that this additional information brings in evident quality improvement on the collected captions. 
Overall, $2,159$ qualified workers annotate $412,690$ valid English captions.

\subsubsection{Chinese Description Collection}
Similar to the English corpus, we collect 10 Chinese descriptions for each video. But to support the video-guided machine translation task, we split these $10$ descriptions into two parts, five directly describing the video content and the other five are the paired translations of $5$ English descriptions for the same video.
All annotations are conducted on the Bytedance Crowdsourcing platform\footnote{A public Chinese crowdsourcing platform: \url{https://zc.bytedance.com}}.
All workers are native Chinese speakers and have a good education background to guarantee that the video content can be correctly understood and the corresponding descriptions can be accurately written. 

\begin{table*}[t]
\small
\setlength{\tabcolsep}{4pt}
\begin{center}
    \begin{tabular}{l c c c r r r r r r r r }
    \toprule
    & & \multicolumn{2}{c}{\textbf{duplicated sent rate}} & \multicolumn{4}{c}{\textbf{\#unique $n$-grams}} & \multicolumn{4}{c}{\textbf{\#unique POS tags}}\\
    \cmidrule(lr){3-4} \cmidrule(lr){5-8} \cmidrule(lr){9-12}
    \textbf{Dataset} & \textbf{sent length} & intra-video & inter-video & 1-gram & 2-gram & 3-gram & 4-gram & verb & noun & adjective & adverb\\
    \midrule
    MSR-VTT     & 9.28 & 66.0\% & 16.5\% & 29,004 & 274,000 & 614,449 & 811,903 & 8,862 & 19,703 & 7,329 & 1,195\\
    \vatexen    & 15.23 & 0 & 0 & 35,589 & 538,517 & 1,660,015 & 2,773,211 & 12,796 & 23,288 & 10,639 & 1,924\\
    \vatexzh    & 13.95 & 0 & 0 & 47,065 & 626,031 & 1,752,085 & 2,687,166 & 20,299 & 30,797 & 4,703 & 3,086  \\
    \bottomrule
    \end{tabular}
\end{center}
\vspace{-1ex}
\caption[Caption for LOF]{We demonstrate the average sentence length, the duplicated sentence rate within a video (intra-video) and within the whole corpus (inter-video), the numbers of unique $n$-grams and POS tags. Our \vatex dataset is lexically richer than MSR-VTT in general. Note that the Chinese POS tagging rules follow the Penn Chinese Treebank standard~\cite{xia2000part}, which is different from English due to different morphemes. For instance, \vatexzh has more nouns and verbs but fewer adjectives than \vatexen, because the semantics of many Chinese adjectives are included in nouns or verbs~\cite{zhang2016sadness}\footnotemark.}
\label{table:vatex-msr}
\end{table*} 

For the first part that directly describes the video content, we follow the same annotation rules as in the collection process of the English captions, except that each Chinese caption must contain at least $15$ Chinese characters.

As for the second part, we aim to collect $5$ English-Chinese parallel pairs for each video to enable the VMT task. However, direct translation by professional translators is costly and time-consuming. Thus, following previous methods~\cite{bouamor2014human,zaghouani2016building} on collecting parallel pairs, we choose the post-editing annotation strategy.
Particularly, for each video, we randomly sample $5$ captions from the annotated $10$ English captions and use multiple translation systems to translate them into Chinese reference sentences. Then the annotation task is,
given the video and the references, the workers are required to post-edit the references and write the parallel Chinese sentence following two rules: (1) the original sentence structure and semantics need be maintained to guarantee the alignment to the corresponding English sentence, and (2) lost or wrong entities and actions could be corrected based on the video content to eliminate the errors from the translation systems.
To further reduce the annotation bias towards one specific translation system, here we use three advanced English$\rightarrow$Chinese translation systems (Google, Microsoft, and self-developed translation systems) to provide the workers with machine-translated sentences as references for each English caption.

In order to ensure the quality of the Chinese captions, we conduct a strict two-stage verification: every collected description must be reviewed and approved by another independent worker. Workers with less than $90\%$ approval rate are blocked.
The interfaces for Chinese caption collection can be found in the supplementary material. Eventually, $450$ Chinese workers participate in these two tasks and write $412,690$ valid Chinese captions. Half of the captions are English-Chinese parallel sentences, so we have $206,345$ translation pairs in total.

\footnotetext{For example, the segmented Chinese word
\begin{CJK*}{UTF8}{gbsn}长发\end{CJK*}
(``long hair") is labeled as one noun in Chinese, but an adjective (``long") and a noun (``hair") in English.}

\begin{figure}
\vspace{-1ex}
    \centering
    \includegraphics[width=0.85\linewidth]{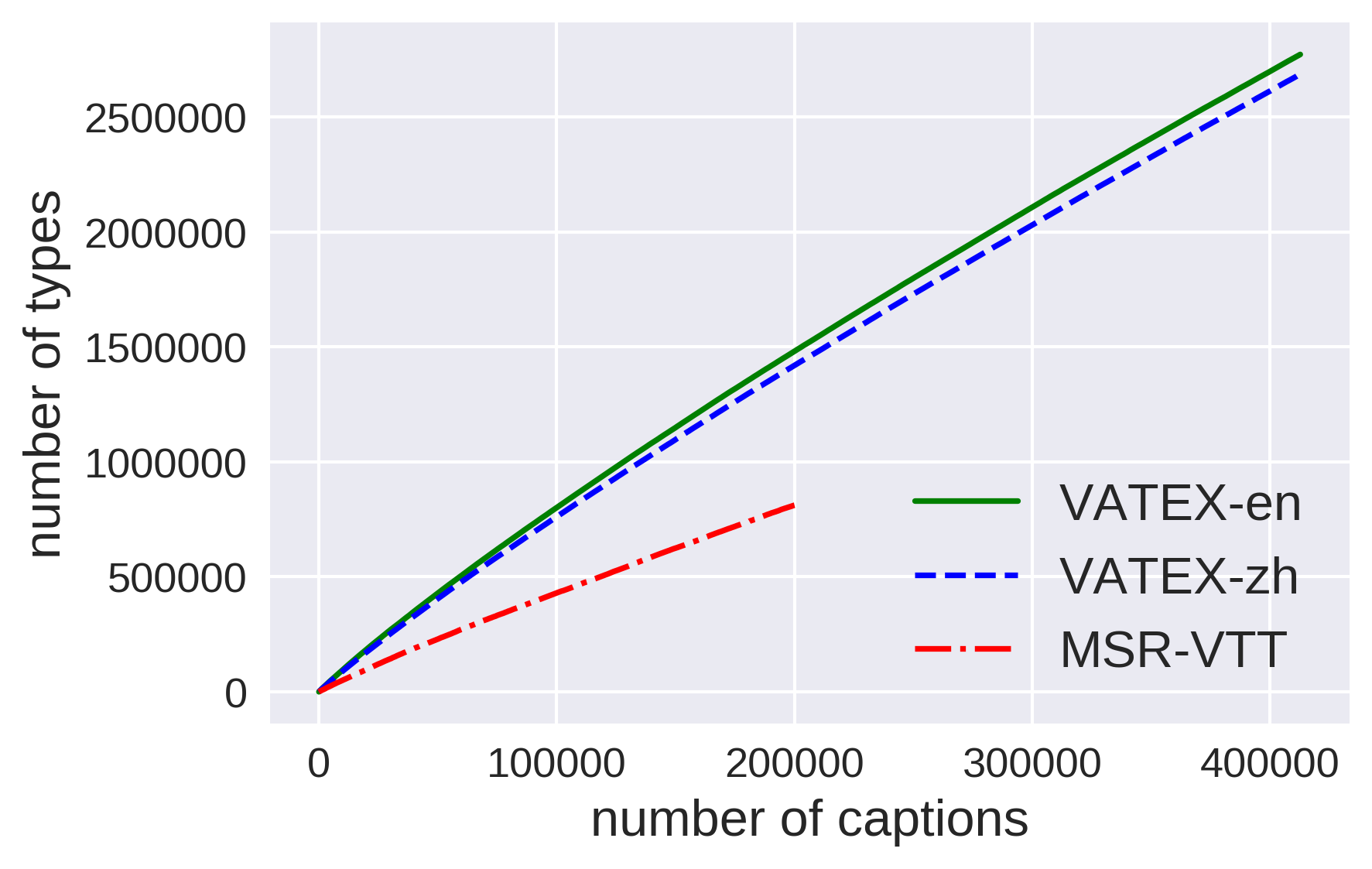}
    \caption{Type-caption curves. Type: unique $4$-gram. \vatex has more lexical styles and caption diversity than MSR-VTT.}
    \label{fig:ngram-cap-curve}
\end{figure}

\subsection{Dataset Analysis}
In Table~\ref{tab:overall}, we briefly compare the overall statistics of the existing video description datasets. In this section, we conduct comprehensive analysis between our \vatex dataset and the MSR-VTT dataset~\cite{msrvtt}, which is the widely-used benchmark for video captioning and the closest to \vatex in terms of domain and scale. Since MSR-VTT only has English corpus, we split \vatex into the English corpus (\vatexen) and the Chinese corpus (\vatexzh) for comparison.
\vatex contains $413$k English and $413$k Chinese captions depicting $41.3$k unique videos from $600$ activities, while MSR-VTT has $200$k captions describing $7$k videos from $257$ activities. 
In addition to the larger scale, the captions in both \vatexen and \vatexzh are longer and more detailed than those in MSR-VTT (see Figure~\ref{fig:cap-per-cap}). 
The average caption lengths of \vatexen, \vatexzh, and MSR-VTT are $15.23$, $13.95$, and $9.28$.

To assess the linguistic complexity, we compare the unique $n$-grams and part-of-speech (POS) tags (\eg, verb, noun, adverb etc.) among MSR-VTT, \vatexen and \vatexzh (see Table~\ref{table:vatex-msr}), which illustrates the improvement of \vatex over MSR-VTT and the difference between the English and Chinese corpora. Evidently, our \vatex datasets represent a wider variety of caption styles and cover a broader range of actions, objects, and visual scenes.

We also perform in-depth comparisons of caption diversity. First, as seen in Table~\ref{table:vatex-msr}, MSR-VTT faces a severe duplication issue in that $66.0\%$ of the videos contains some exactly same captions, while our \vatex datasets are free of this problem and guarantee that the captions within the same video are unique. Not only within videos, but the captions in our \vatex datasets are also much more diverse even within the whole corpus, which indicates that our \vatex can also be a high-quality benchmark for video retrieval.

For a more intuitive measure of the lexical richness and caption diversity, we then propose the \emph{Type-Caption Curve}, which is adapted from the type-token vocabulary curve~\cite{type-token} but specially designed for the caption corpora here. The total number of captions and the number of distinct vocabulary words (types) are computed for each corpus. So we plot the number of types against the number of captions for MSR-VTT, \vatexen, and \vatexzh (see Figure~\ref{fig:ngram-cap-curve} where we choose $4$-grams as the types).
From these type-caption curves, inferences are drawn about lexical style or caption diversity (vocabulary use), as well as lexical competence (vocabulary size), so our \vatex datasets are shown to be more linguistically complex and diverse.

\section{\vatex Tasks}
\subsection{Multilingual Video Captioning}
\label{subsec:ml-cap}
Multilingual video captioning is the task of describing the content of a video using more than one language such as English and Chinese. Below we first introduce a baseline model for monolingual video captioning and then present three different models for multilingual video captioning. 

\begin{figure}
    \centering
    \includegraphics[width=0.45\textwidth]{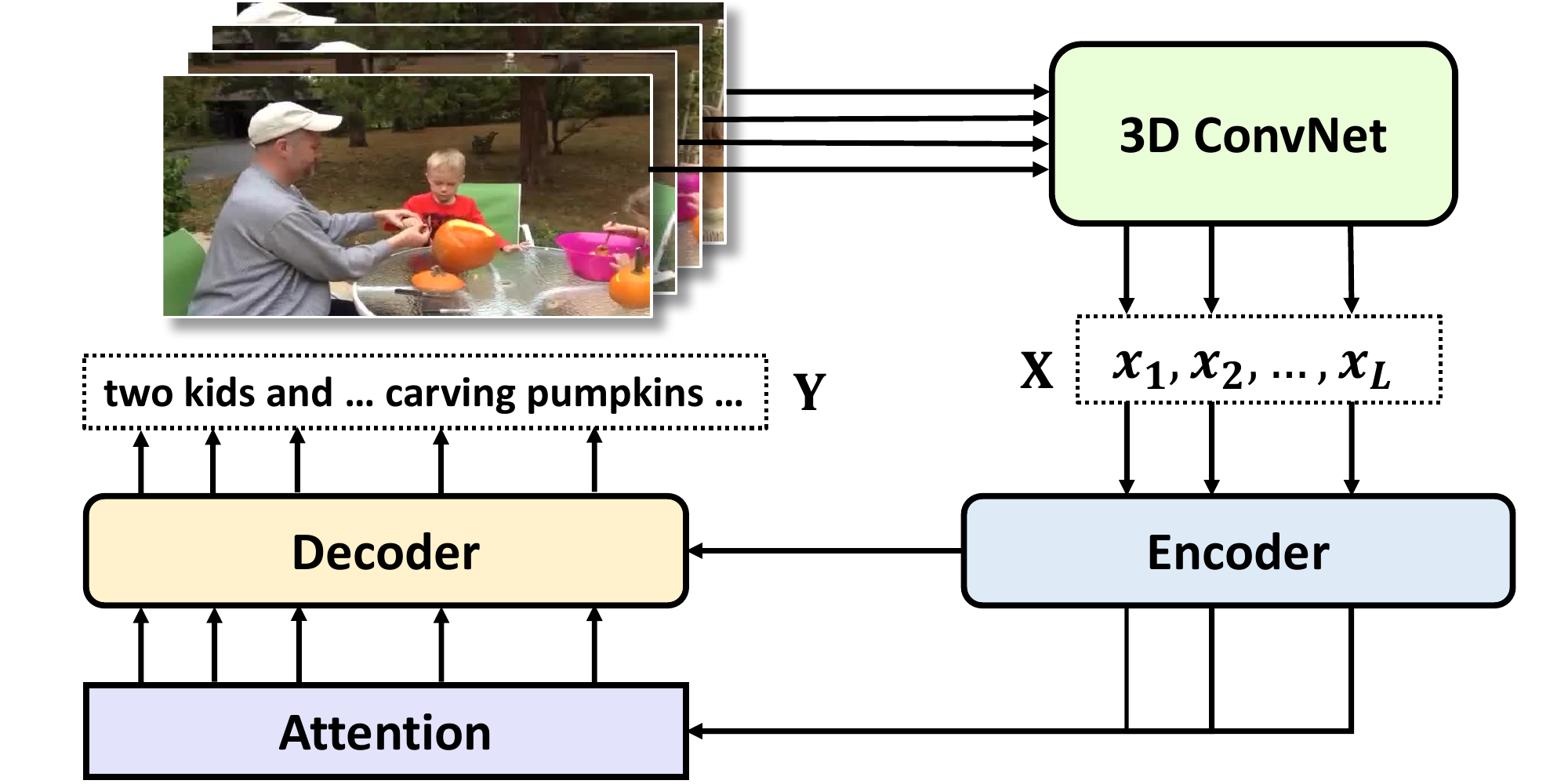}
    \caption{Monolingual video captioning model.}
    \label{fig:cap_model}
\end{figure}

\subsubsection{Models}
We begin with the well-known attention-based encoder-decoder model for video captioning. As illustrated in Figure~\ref{fig:cap_model}, there are three main modules to this architecture:
\begin{itemize}
    \item A 3D convolutional neural network (3D ConvNet) that learns the spatiotemporal features of the video and outputs a sequence of segment-level features $X=\{x_1, x_2, \dots, x_L\}$.
    \item A video encoder module $f_{enc}$ that encodes $X$ into video-level features $V = \{v_1, v_2, \dots, v_L\}$ by modeling long-range temporal contexts.
    \item An attention-based language decoder module $f_{dec}$ that produces a word $y_t$ at every time step $t$ by  considering the word at previous step $y_{t-1}$, the visual context vector $c_t$ learned from the attention mechanism.
\end{itemize}
We instantiate the captioning model by adapting the model architectures from the state-of-the-art video captioning methods~\cite{Pasunuru2017ReinforcedVC,wang2018video}. We employ the pretrained I3D model~\cite{I3D} for action recognition as the 3D ConvNet to obtain the visual features $X$, Bidirectional LSTM~\cite{Schuster:1997:BiRNN} (bi-LSTM) as the video encoder $f_{enc}$, and LSTM~\cite{LSTM} as the language decoder $f_{dec}$. We also adopt the dot-product attention, so at the decoding step $t$, we have 
\vspace{-1ex}
\begin{equation}
    y_t, h_t = f_{dec}([y_{t-1}, c_t], h_{t-1}) ~,
\end{equation}
where $h_t$ is the hidden state of the decoder at step $t$ and 
\begin{align}
\label{equ:temporal-att}
    c_t &= \text{softmax}(h_{t-1} W V^T)V ~,
\end{align}
where $W$ is a learnable projection matrix.

\begin{figure}
    \centering
    \includegraphics[width=0.4\textwidth]{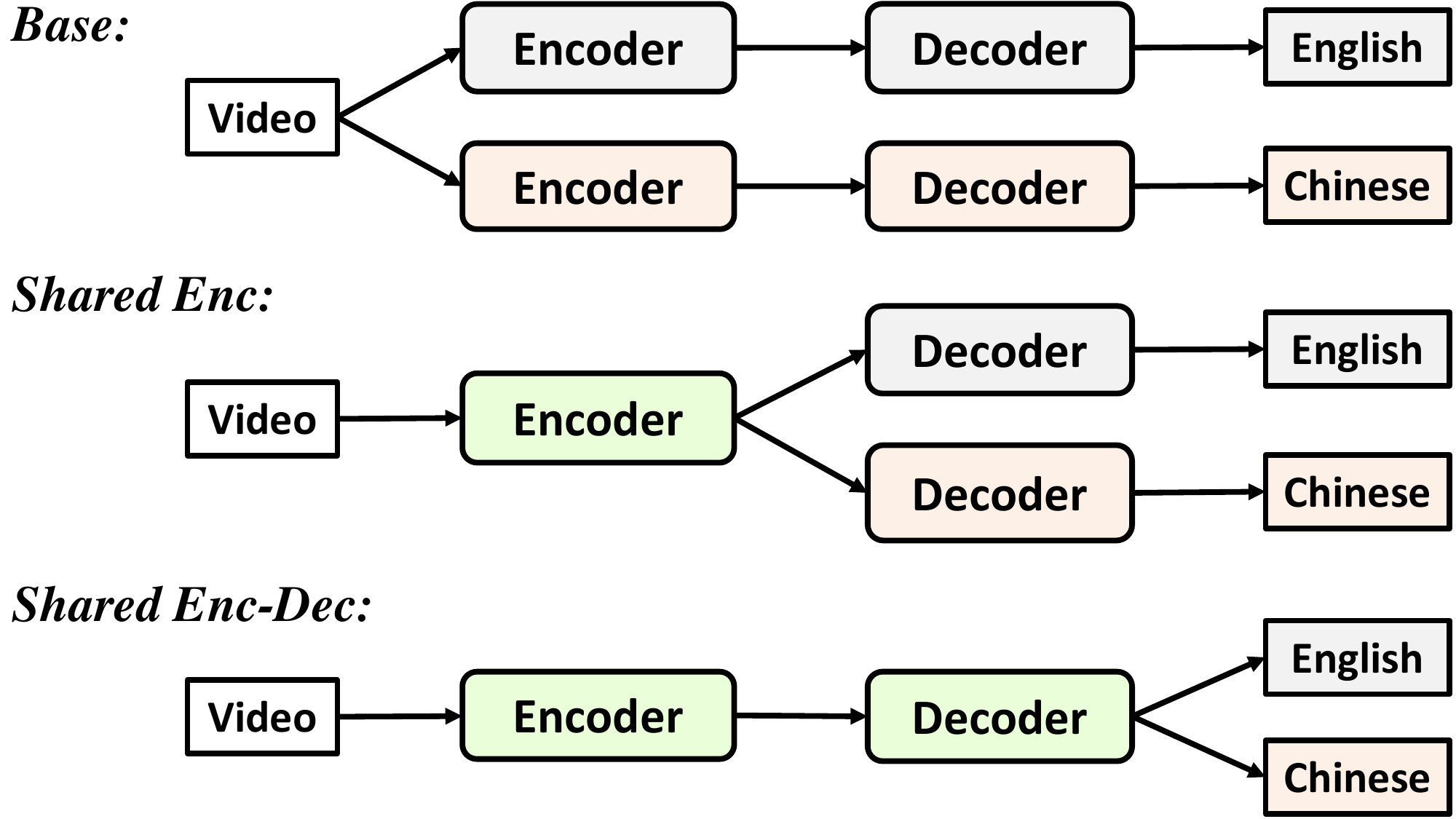}
    \caption{Multilingual video captioning models.}
    \label{fig:ml_cap_model}
\end{figure}

To enable multilingual video captioning, we examine three methods (see Figure~\ref{fig:ml_cap_model}): (1) Two \textbf{Base} models, which are two monolingual encoder-decoder models (as described in Figure~\ref{fig:cap_model}) trained separately for either English or Chinese; (2) A \textbf{Shared Enc} model, which has a shared video encoder but two language decoders to generate English and Chinese; (3) A \textbf{Shared Enc-Dec} model, where there are just one encoder and one decoder, both shared by English and Chinese, and the only difference is that the word embedding weight matrices are different for different languages.

\begin{table*}[t]
\begin{center}
\small
\resizebox{\textwidth}{!}{
  \begin{tabular}{ l c | cccc | cccc }
      \toprule
      &  &  \multicolumn{4}{c}{\textbf{English}} &\multicolumn{4}{|c}{\textbf{Chinese}} \\
      \cmidrule(lr){3-6}\cmidrule(lr){7-10}
        \textbf{Model} & \textbf{\#Params}
         & BLEU-4 & Meteor & Rouge-L & CIDEr
         & BLEU-4 & Meteor & Rouge-L & CIDEr \\
        \midrule
        Base w/o WT & 52.5M &
        \confnum{28.1}{0.38} & \confnum{\textbf{21.7}}{0.15} & \confnum{46.8}{0.18} & \confnum{44.3}{0.98} &
        \confnum{24.4}{0.86} & \confnum{29.6}{0.30} & \confnum{51.3}{0.43} & \confnum{34.0}{0.11} \\
        Base & 39.7M &
        \confnum{28.1}{0.32} & \confnum{21.6}{0.19} & \confnum{46.9}{0.16} & \confnum{44.3}{0.10} &
        \confnum{\textbf{24.9}}{0.20} & \confnum{29.7}{0.21} & \confnum{51.5}{0.28} & \confnum{34.7}{0.47} \\
        Shared Enc & 34.9M &
        \confnum{\textbf{28.4}}{0.21} & \confnum{\textbf{21.7}}{0.65} & \confnum{\textbf{47.0}}{0.09} & \confnum{\textbf{45.1}}{0.25} & 
        \confnum{\textbf{24.9}}{0.26} & \confnum{29.7}{0.11} & \confnum{51.6}{0.20} & \confnum{34.9}{0.40} \\
        Shared Enc-Dec  & \textbf{26.3M} &
        \confnum{27.9}{0.50} & \confnum{21.6}{0.55} & \confnum{46.8}{0.19} & \confnum{44.2}{0.23} & 
        \confnum{\textbf{24.9}}{0.25} & \confnum{\textbf{29.8}}{0.23} & \confnum{\textbf{51.7}}{0.09} & \confnum{\textbf{35.0}}{0.18} \\
        \bottomrule
  \end{tabular}
}
\end{center}
\vspace{-1ex}
\caption{Multilingual video captioning.
We report the results of the baseline models in terms of BLEU-4, Meteor, and Rouge-L, and CIDEr scores. \textit{Each model is trained for five times with different random seeds and the results are reported with a confidence level of 95\%}. WT: weight tying, which means the input word embedding layer and the softmax layer share the same weight matrix.}
\label{table:ml-cap}
\end{table*} 

\subsubsection{Experimental Setup}
\label{subsubsec:setup}
\noindent\textbf{Implementation Details.}
We train the models on the \vatex dataset following the splits in Table~\ref{table:split}. To preprocess the videos, we sample each video at $25fps$ and extract the I3D features~\cite{I3D} from these sampled frames. The I3D model is pretrained on the original Kinetics training dataset~\cite{kay2017kinetics} and used here without fine-tuning. More details about data preprocessing and implementation can be found in the supplementary material.

\noindent\textbf{Evaluation Metrics.}
We adopt four diverse automatic evaluation metrics: BLEU~\cite{Papineni2001BleuAM}, Meteor~\cite{Denkowski2014MeteorUL}, Rouge-L~\cite{Lin2004ROUGEAP}, and CIDEr~\cite{Vedantam2015CIDErCI}. We use the standard evaluation code from MS-COCO server~\cite{Chen2015MicrosoftCC} to obtain the results. 

\subsubsection{Results and Analysis}
Table~\ref{table:ml-cap} shows the results of the three baseline models on both English and Chinese test sets. 
The performances of the multilingual models (\textit{Shared Enc} and \textit{Shared Enc-Dec}) are consistently (though not significantly) improved over the monolingual model (\textit{Base}). It indicates that multilingual learning indeed helps video understanding by sharing the video encoder. 
More importantly, the parameters of the \textit{Shared Enc} and \textit{Shared Enc-Dec} are significantly reduced by $4.7$M and $13.4$M over the \textit{Base} models. 
These observations validate that a compact unified model is able to produce captions in multiple languages and benefits from multilingual knowledge learning.
We believe that more specialized multilingual models would improve the understanding of the videos and lead to better results. Furthermore, incorporating multimodal features like audio~\cite{wang2018watch} would further improve the performance, which we leave for future study. 

\subsection{Video-guided Machine Translation}
In this section, we discuss the enabled new task, Video-guided Machine Translation (VMT), to translate a source language sentence into the target language using the video information as additional spatiotemporal context.
This task has various potential real-world applications, e.g., translating posts with the video content in social media.

\subsubsection{Method}
\begin{figure}
    \centering
    \includegraphics[width=0.9\linewidth]{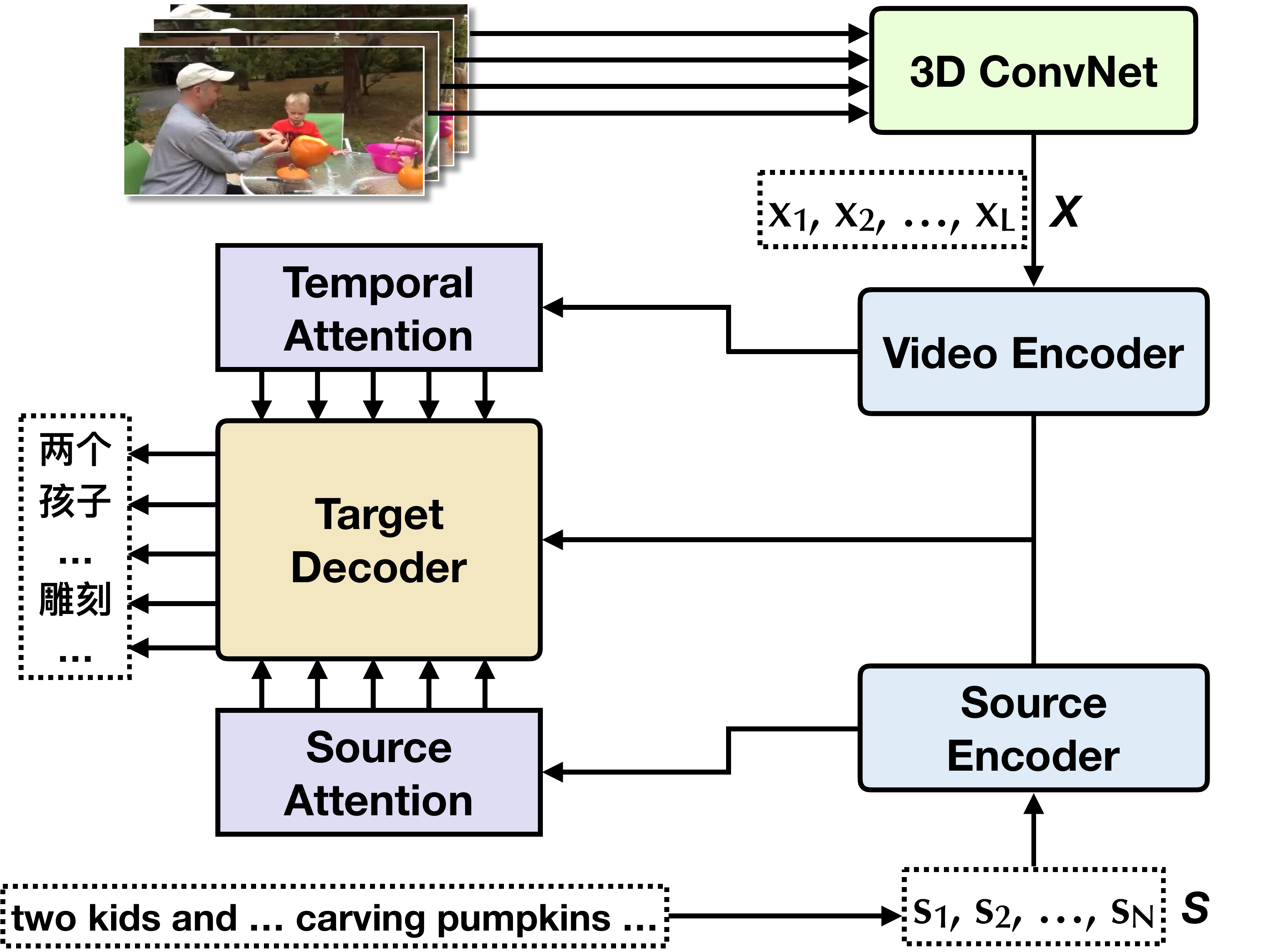}
    \caption{Video-guided machine translation model.}
    \label{fig:mmt_model}
\end{figure}

In VMT, the translation system takes a source sentence and the corresponding video as the input, and generates the translated target sentence. 
To effectively utilize the two modalities, text and video, we design a multimodal sequence-to-sequence model~\cite{sutskever2014sequence,venugopalan2015sequence} with the attention mechanism~\cite{bahdanau2014neural,luong2015effective} for VMT. 
The overview of our model is shown in Figure~\ref{fig:mmt_model}, which mainly consists of the following three modules.

\noindent\textbf{Source Encoder.} 
For each source sentence represented as a sequence of $N$ word embeddings $S=\{s_1, s_2, \dots, s_N\}$, the source encoder $f_{enc}^{src}$ transforms it into the sentence features $U =\{u_1, u_2, \dots, u_N\}$.

\noindent\textbf{Video Encoder.}
Similar in Section~\ref{subsec:ml-cap}, we use a 3D ConvNet to convert each video into a sequence of segment-level features $X$. Then we employ a video encoder $f_{enc}^{vi}$ to transform $X$ into the video features $V=\{v_1, v_2, \dots, v_L\}$.

\noindent\textbf{Target Decoder.}
The sentence embedding from the source language encoder $f_{enc}^{src}$ and the video embedding from the video encoder $f_{enc}^{vi}$ are concatenated and fed into the target language decoder $f_{dec}^{tgt}$.
To dynamically highlight the important words of the source sentence and the crucial spatiotemporal segments in the video, we equip the target decoder $f_{dec}^{src}$ with two attention mechanisms. Thus, at each decoding step $t$, we have
\begin{equation}
    y_t, h_t = f_{dec}^{tgt}([y_{t-1}, c^{src}_t, c^{vi}_t], h_{t-1}) ~,
\end{equation}
where $h_t$ is the hidden state of the decoder at step $t$. $c^{vi}_t$ is the video context vector that is computed with the temporal attention of the video segments (see Equation~\ref{equ:temporal-att}), and $c^{src}_t$ is the source language context vector:
\begin{equation}
    c^{src}_t = \text{softmax}(h_{t-1} W^{src} U^T) U ~,
\end{equation}
where $W^{src}$ is a learnable projection matrix.

\subsubsection{Experimental Setup}
\noindent\textbf{Baselines.}
We consider the following three baselines to compare: 
(1) \textit{Base NMT Model}: We only consider the text information for machine translation and adopt the encoder-decoder model with the source attention mechanism. 
(2) \textit{Average Video Features}: We average the segment-level features $X$ of each video as $\overline{x}$. The average video feature $\overline{x}$ is then concatenated with each word embedding $s_t$ in $S$. The model structure is the same as the base NMT model. 
(3) \textit{LSTM Video Features}: This is our VMT model without the temporal attention for videos in the decoder.


\begin{table}[t]
\setlength{\tabcolsep}{4pt}
\begin{center}
\resizebox{0.48\textwidth}{!}{
    \begin{tabular}{lcc}
    \toprule
    \textbf{Model} & \textbf{English}$\rightarrow$\textbf{Chinese} & \textbf{Chinese}$\rightarrow$\textbf{English} \\
    \midrule
    Base NMT w/o VI & 26.85~\en & 24.31~\en \\
    $+$ Average VI & 26.97~\up{0.12} & 24.39~\up{0.08} \\
    $+$ LSTM VI w/o Attn & 27.43~\up{0.58} & 24.76~\up{0.45} \\
    $+$ LSTM VI w/ Attn (VMT) & \textbf{29.12}~\up{2.27} & \textbf{26.42}~\up{2.11} \\
    \bottomrule
    \end{tabular}
}
\end{center}
\vspace{-2ex}
\caption{
Video-guided Machine Translation. Results are reported on the BLEU-4 scores. VI: video features from the pretrained I3D model. Attn: temporal attention mechanism.}
\label{table:mmt}
\end{table}

\subsubsection{Results and Analysis}
\noindent\textbf{VMT.} 
We first show in Table~\ref{table:mmt} the results of four different models on \textit{Chinese$\rightarrow$English} and \textit{English$\rightarrow$Chinese} translations. 
The marginal improvements by the \textit{Average video Features} and the \textit{LSTM Video Features} reveal that, passively receiving and incorporating the video features is ineffective in helping align source and target languages. 
However, we can observe that the translation system achieves much better performance when using the \textit{LSTM Video Features} with temporal attention (our full VMT model) to dynamically interact with the video features. 
It is because that with the attention mechanism, the language dynamics are used as a query to highlight the relevant spatiotemporal features in the video, and then the learned video context would assist the word mapping between source and target language spaces. This also validates that extra video information can be effectively utilized to boost machine translation systems. 

\noindent\textbf{Masked VMT.}
Videos contain rich information on subject/object nouns and action verbs. Therefore, we conduct noun/verb masking experiments~\cite{Caglayan2019Probing} to investigate to what extent the video information can help machine translation. 
We randomly replace $0\%/25\%/50\%/75\%/100\%$ nouns or verbs of the English captions with a special token [M], and then train the NMT and VMT models on the \textit{Chinese$\rightarrow$English} translation task with different masking rates. This experimental design is to evaluate the capability of VMT in recovering the missing information of the source sentence with the help of the video context.  

In addition to the BLEU-4 metric, we propose to use the noun/verb recovery accuracy, which is the percentages of the correctly translated nouns/verbs in the target sentences, to precisely evaluate the impact of additional video information on recovering nouns/verbs. 
The results with different masking rates are shown in Table~\ref{table:mmt-analysis}.   
First, the VMT model consistently outperforms the NMT model with different masking rates on both metrics.
Moreover, as the masking rate increases, the NMT model struggles to figure out the correct nouns/verbs because of the scarce parallel caption pairs; while the VMT model can rely on the video context to obtain more useful information for translation, and thus the performance gap on the recovery accuracy increases dramatically. It shows that in our VMT model, video information can play a crucial role in understanding subjects, objects, and actions, as well as their relations.

\begin{table}[t]
\setlength{\tabcolsep}{3pt}
\begin{center}
\resizebox{0.48\textwidth}{!}{
    \begin{tabular}{lcccccccccc}
    \toprule
    & \multicolumn{5}{c}{\textbf{BLEU-4}} & \multicolumn{5}{c}{\textbf{Accuracy (\%)}} \\
    \cmidrule(lr){2-6} \cmidrule(lr){7-11}
    \textbf{Rate} & $0\%$ & $25\%$ & $50\%$ & $75\%$ & $100\%$ & $0\%$ & $25\%$ & $50\%$ & $75\%$ & $100\%$ \\
    \midrule
    \multicolumn{11}{c}{Noun Masking} \\
    \midrule
    NMT & 26.9 & 20.2 & 13.0 & 8.5 & 4.1 & 70.2 & 53.7 & 35.4 & 15.6 & 10.1 \\
    VMT & 29.1 & 24.7 & 19.3 & 16.9 & 14.3 & 76.4 & 65.6 & 50.8 & 43.2 & 39.7 \\
    \midrule
    \multicolumn{11}{c}{Verb Masking} \\
    \midrule
    NMT & 26.9 & 23.3 & 15.4 & 11.6 & 7.2 & 65.1 & 57.4 & 40.9 & 33.6 & 19.8 \\
    VMT & 29.1 & 26.8 & 22.0 & 19.3 & 16.5 & 70.4 & 63.6 & 54.2 & 48.7 & 40.5 \\
    \bottomrule
    \end{tabular}
}
\end{center}
\vspace{-2ex}
\caption{Video-guided machine translation on \textit{English$\rightarrow$Chinese} with different noun/verb masking rates. We evaluate the results using the BLEU-4 score and noun/verb recovery accuracy.}
\label{table:mmt-analysis}
\end{table}

\section{Discussion and Future Work}
In this paper, we introduce a new large-scale multilingual dataset for video-and-language research. 
In addition to (multilingual) video captioning and video-guided machine translation, there are also some other potentials of this dataset. For example, since the natural language descriptions in \vatex are unique, one promising direction is to use multilingual descriptions of our dataset as queries to retrieve the video clip from all videos~\cite{Lin2014VisualSS} or even localize it within an untrimmed long video~\cite{zhang2019man}. 
Meanwhile, \vatex has 600 fine-grained action labels, so we can holdout certain action classes to evaluate the generalizability of different video captioning models to support zero-/few-shot learning~\cite{wang2019learning}. 
Furthermore, our dataset can contribute to other research fields like Neuroscience. For instance, when describing the same videos, the focus points of people using different languages can be reflected by their written captions. By analyzing multilingual captions, one can likely infer the commonality and discrepancy on the brain attention of people with different cultural and linguistic backgrounds.
In general, we hope the release of our \vatex dataset would facilitate the advance of video-and-language research.

{\small
\bibliographystyle{ieee_fullname}
\bibliography{egbib}
}

\clearpage

\appendix

\section*{Supplementary Material}

\section{Implementation Details}
\paragraph{Multilingual Video Captioning}
To preprocess the videos, we sample each video at $25fps$ and extract the I3D features~\cite{I3D} from these sampled frames. The I3D model is pretrained on the original Kinetics training dataset~\cite{kay2017kinetics} and used here without fine-tuning.
Both the English and Chinese captions are truncated to a maximum of 30 words. 
Note that we use the segmented Chinese words\footnote{We use the open-source tool Jieba for Chinese word segmentation: \url{https://github.com/fxsjy/jieba}} rather than raw Chinese characters. The vocabularies are built with a minimum word count $5$, resulting in around $11,000$ English words and about $14,000$ Chinese words.

All the hyperparameters are tuned on the validation sets and same for both English and Chinese caption training. The video encoder is a bi-LSTM of size $512$ and the decoder LSTM is of size $1024$. The dimensions of the word embedding layers are $512$. 
All models are trained using MLE loss and optimized using Adam optimizer~\cite{kingma2014adam} with a batch size $256$. We adopt Dropout for regularization. The learning rate is initially set as $0.001$ and then halted when the current CIDEr score does not surpass the previous best for $4$ epochs. Schedule sampling~\cite{Bengio2015ScheduledSF} is employed to train the models. The probability of schedule sampling is first set to be 0.05, then increased
by 0.05 every 5 epochs, and eventually fixed at 0.25 after 25 epoches. At test time, we use beam search of size $5$ to report the final results.

\paragraph{Video-guided Machine Translation}
The data prepossessing steps are the same as above except that we truncate the captions with a maximum length of $40$ here. The baseline NMT is composed of a $2$-layer bi-LSTM encoder of size $512$ and a $2$-layer LSTM decoder of size $1024$. The dimensions of both English and Chinese word embeddings are $512$. The video encoder is a bi-LSTM of size $512$. MLE loss is implemented to train the model using Adam optimizer~\cite{kingma2014adam}. The batch size is $32$ during training and early-stopping is used to choose the models. Then we fine-tune the private parameters of model for each language with the shared parameters fixed. As for evaluation, we use beam search of size $5$ to report the results on the BLEU-4 metric.


\section{Data Collection Interfaces}
\begin{figure*}
    \centering
    \includegraphics[width=\textwidth]{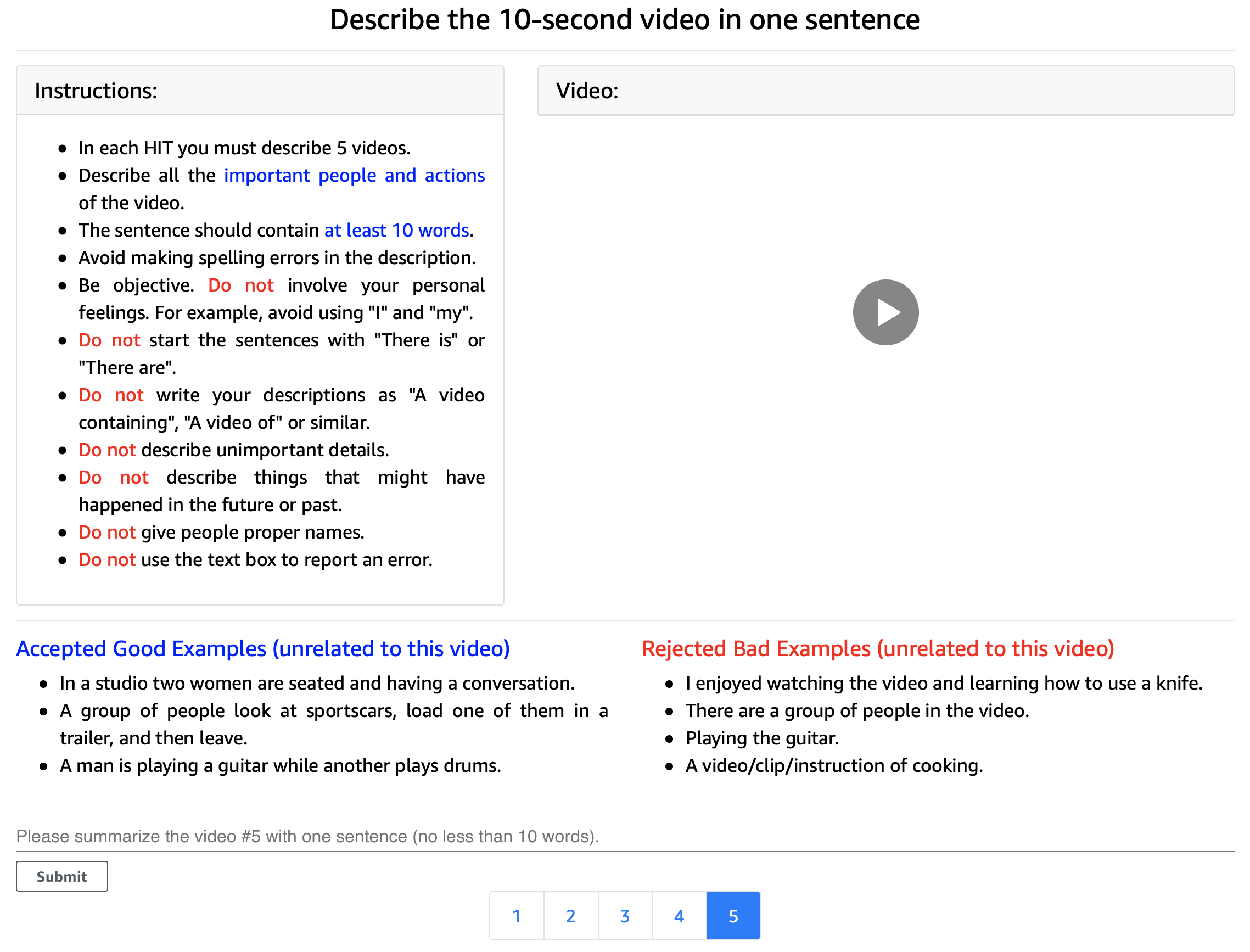}
    \caption{The AMT interface for collecting the English captions. In each assignment, the workers are required to annotate $5$ video clips. The instructions are kept visible for each clip. We provide the workers with the accepted good examples and rejected bad examples to further improve the quality of annotations. Note that the given examples are unrelated to the current video clips.}
    \label{fig:en_interface}
\end{figure*}

\begin{figure*}
    \centering
    \includegraphics[width=\textwidth]{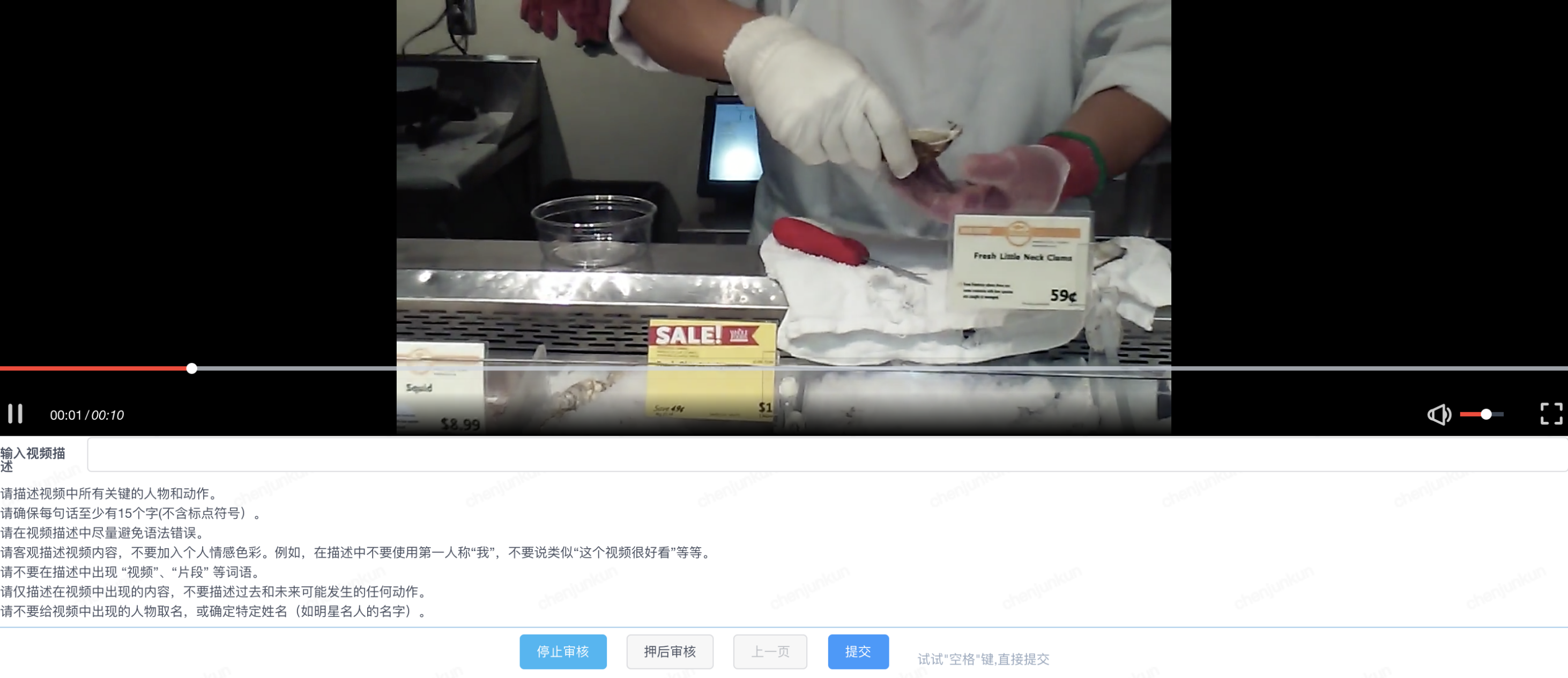}
    \caption{The interface for collecting the Chinese captions by directly describing the video content. In each assignment, the workers are required to annotate $1$ video clip. The instructions are kept visible for each clip. After the first-stage annotation, each Chinese caption must be reviewed and approved by another independent worker.}
    \label{fig:zh_interface_video}
\end{figure*}

\begin{figure*}
    \centering
    \includegraphics[width=\textwidth]{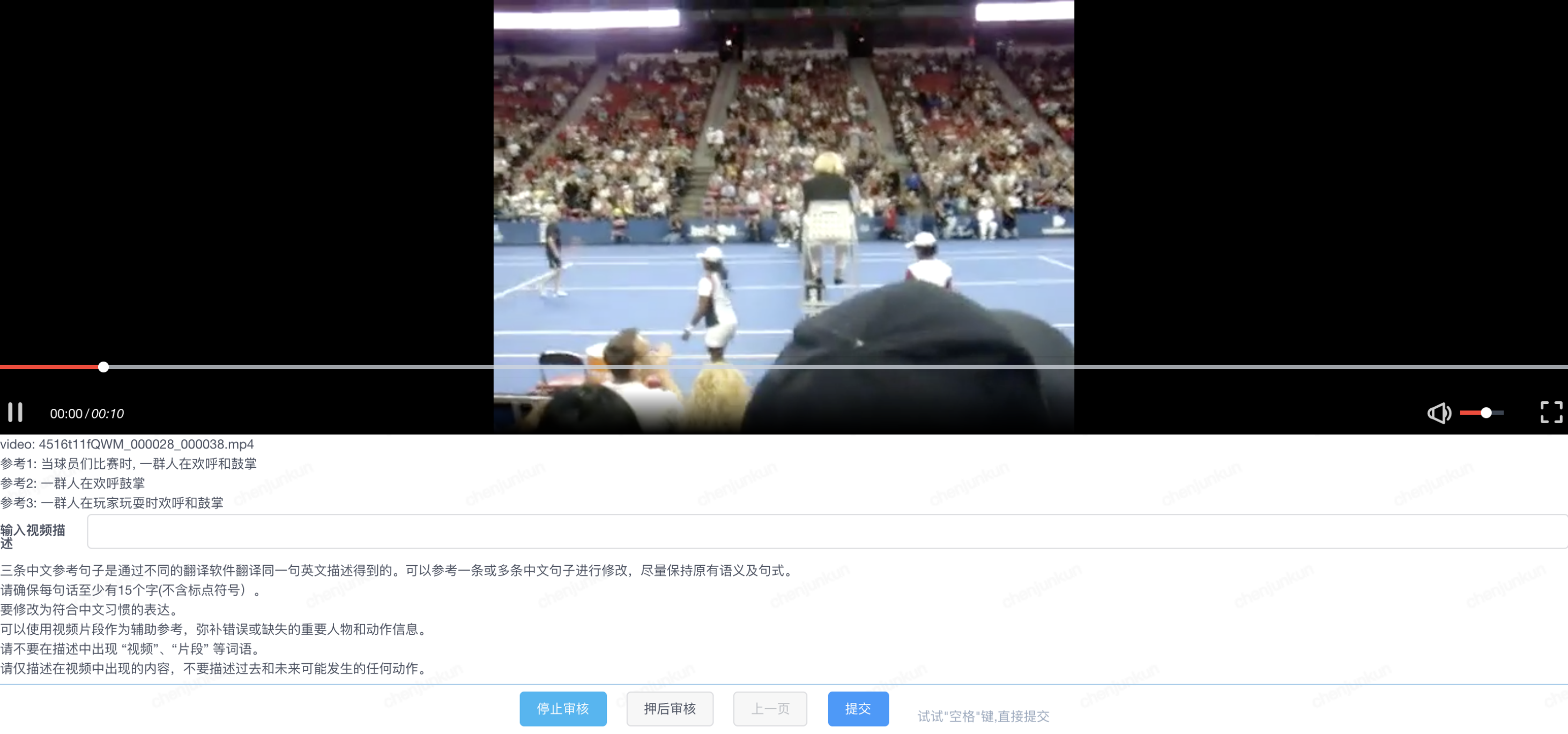}
    \caption{The interface for collecting the Chinese captions by post-editing the translated reference sentences and watching the video clips. In each assignment, the workers are required to annotate $1$ video clip. The instructions are kept visible for each clip. We provide the workers with three reference sentences translated by Google, Microsoft and Self-developed translation systems. Note that the order of three reference sentences is randomly shuffled for each video clip to reduce the annotation bias towards one specific translation system. After the first-stage annotation, each Chinese caption must be reviewed and approved by another independent worker.}
    \label{fig:zh_interface_en}
\end{figure*}

We show the AMT interface for English caption collection in Figure~\ref{fig:en_interface}. Since the Chinese captions are divided into two parts, we build two separate interfaces, one of which is to collect the captions that directly describe the video (Figure~\ref{fig:zh_interface_video}) and the other for collecting the Chinese translations parallel to the English captions (Figure~\ref{fig:zh_interface_en}).

\section{More \vatex Samples}
\begin{figure*}
    \centering
    \includegraphics[width=\textwidth]{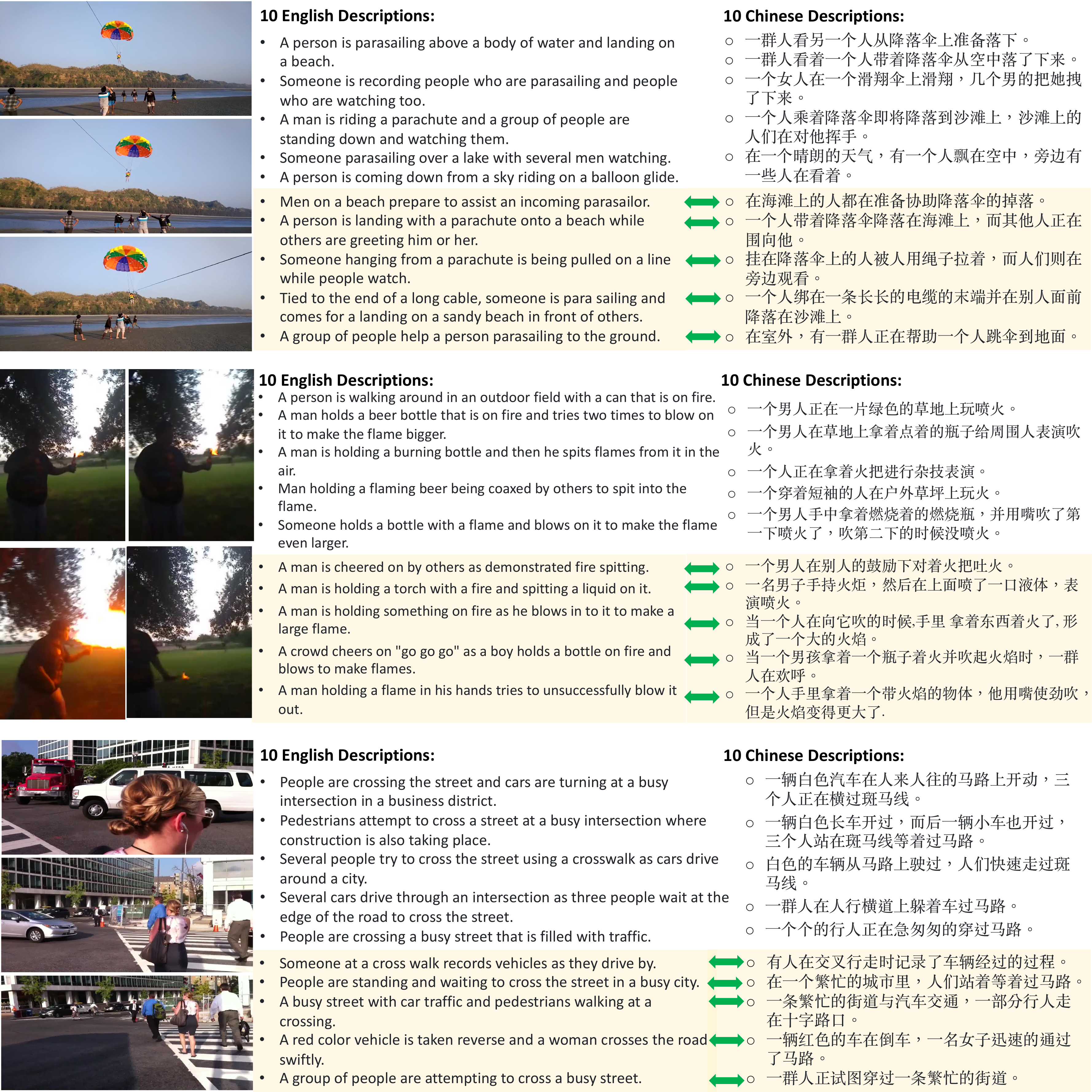}
    \caption{More samples of our \vatex dataset. Each video has $10$ English and $10$ Chinese descriptions. All depicts the same video and thus are distantly parallel to each other, while the last five are the paired translations to each other.}
    \label{fig:sample_supp}
\end{figure*}

In addition to the example shown in the main paper, Figure~\ref{fig:sample_supp} demonstrates more samples of our \vatex dataset. 

\section{Qualitative Results}
\begin{figure*}
    \centering
    \includegraphics[width=\textwidth]{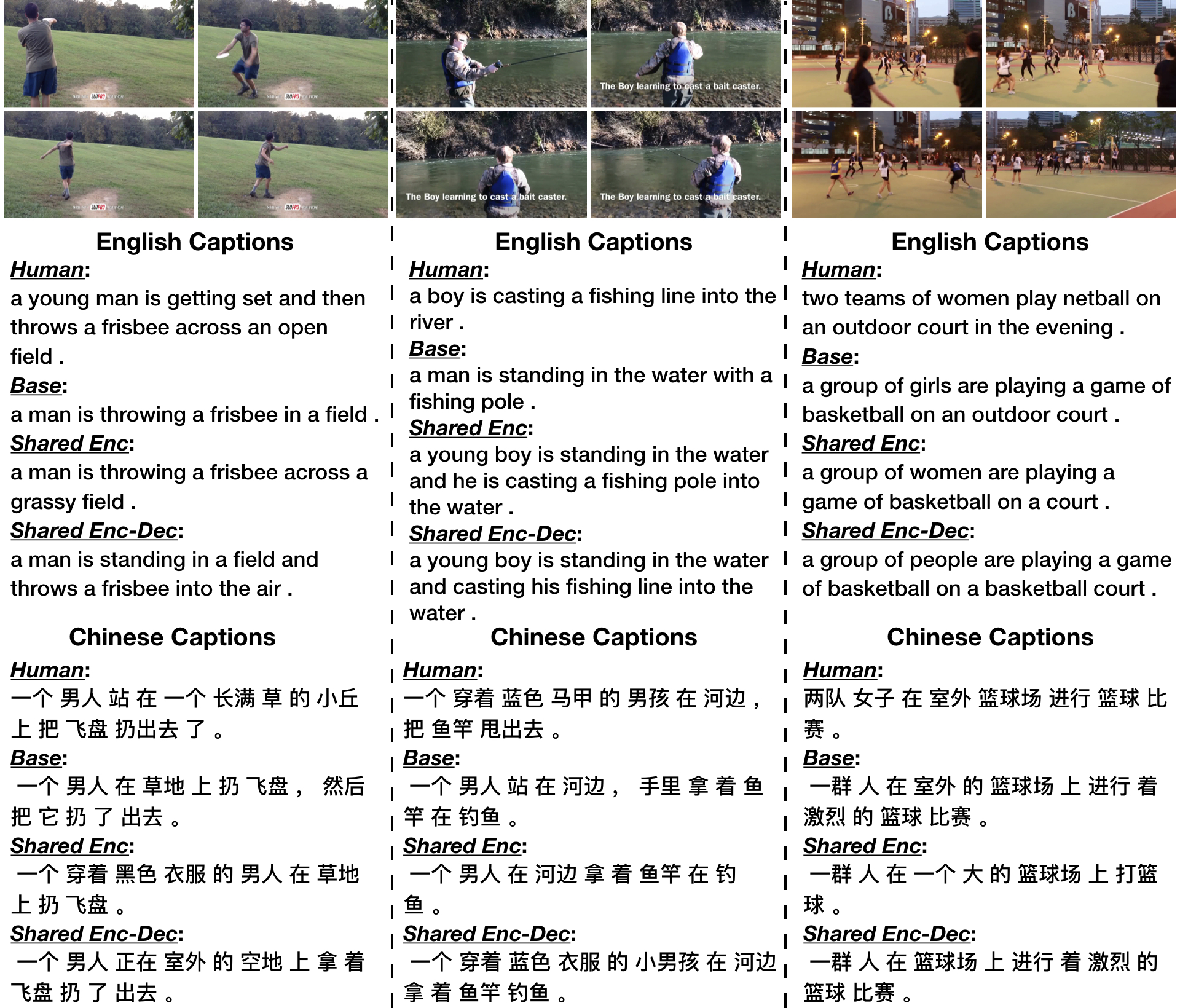}
    \caption{Qualitative comparison among different methods of multilingual video captioning on the \vatex dataset. Both the English and Chinese results are shown. For each video sample, we list a human-annotated caption and the generated results by three models, \textit{Base}, \textit{Shared Enc}, and \textit{Shared Enc-Dec}. The multilingual models (\textit{Shared Enc} and \textit{Shared Enc-Dec}) can generate more coherent and informative captions than the monolingual model (\textit{Base}).}
    \label{fig:captioning_supp}
\end{figure*}

\paragraph{Multilingual Video Captioning}
Figure~\ref{fig:captioning_supp} illustrates some qualitative examples of multilingual video captioning, where we compare both the English and Chinese results generated by the monolingual models (\textit{Base}), the multilingual model that shares the video encoder for English \& Chinese (\textit{Shared Enc}), and the multilingual model that shares both the video encoder and the language decoder for English \& Chinese (\textit{Shared Enc-Dec}).  

\begin{figure*}
    \centering
    \includegraphics[width=\textwidth]{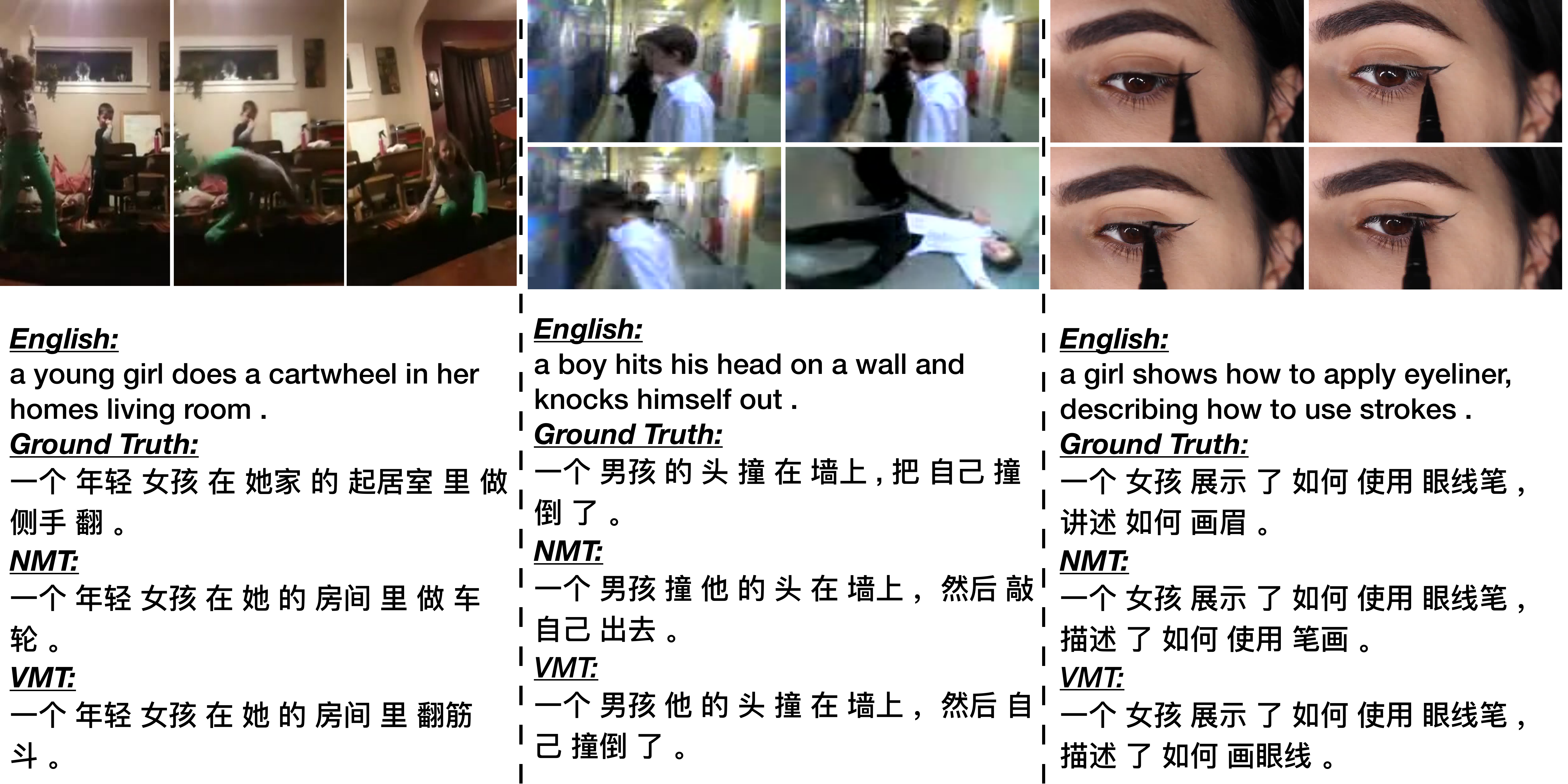}
    \caption{Qualitative comparison between neural machine translation (NMT) and video-guided machine translation (VMT) on the \vatex dataset. For each video sample, we list the original English description and the translated sentences by the base NMT model and our VMT model. The NMT model mistakenly interprets some words and phrases, while the VMT model can generate more precise translation with the corresponding video context.}
    \label{fig:vmt_supp}
\end{figure*}

\begin{figure*}
    \centering
    \includegraphics[width=\textwidth]{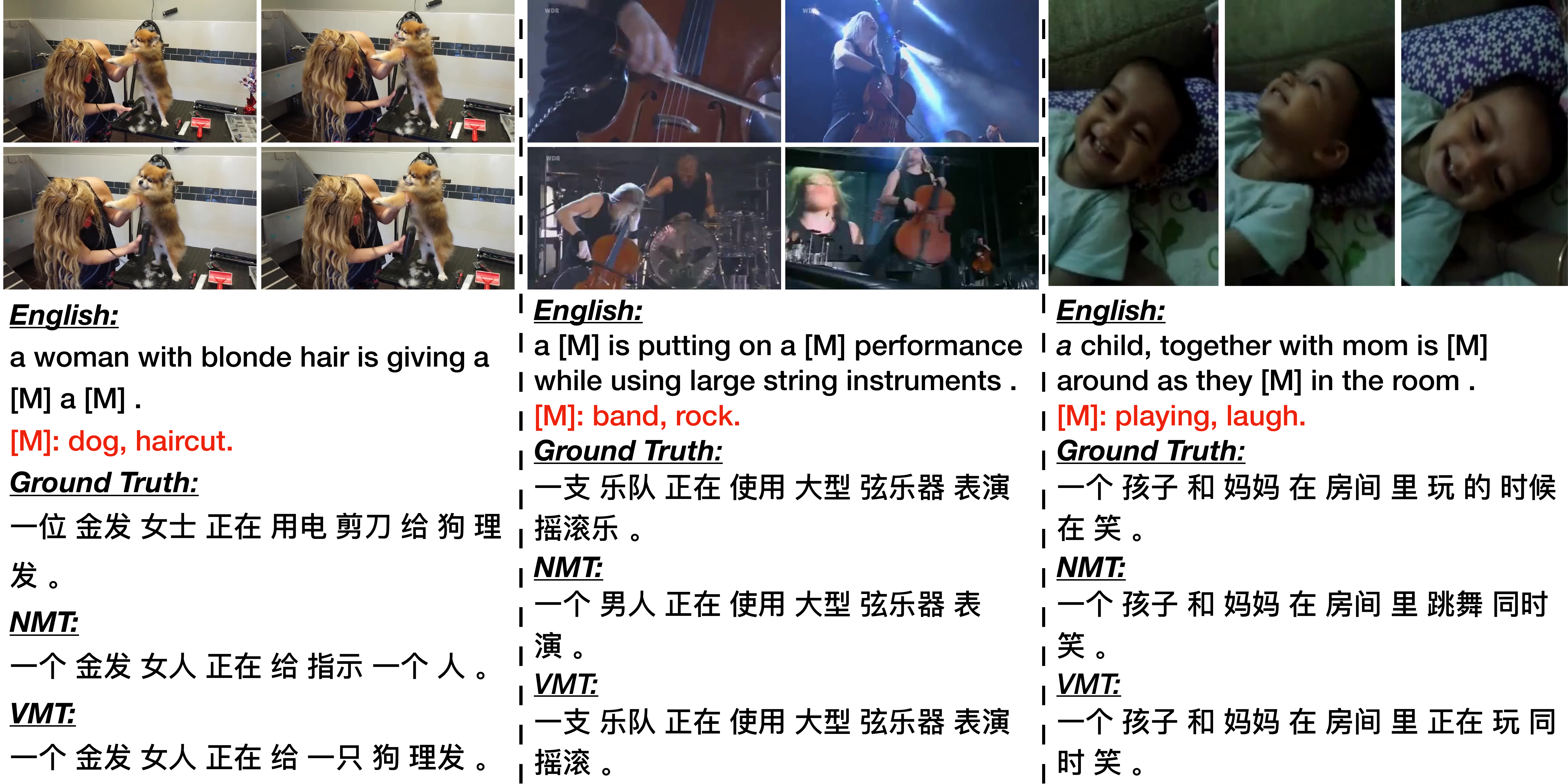}
    \caption{Qualitative comparison between masked neural machine translation (NMT) and masked video-guided machine translation (VMT) on the \vatex dataset. The nouns/verbs in English captions are randomly replaced by a special token [M]. For each video sample, we list the original English description and the translated sentences by the base NMT model and our VMT model. The NMT model struggles to figure out the correct nouns/verbs because of the scarce parallel pairs, while the VMT model can rely on the video context to recover the masked nouns/verbs.}
    \label{fig:mvmt_supp}
\end{figure*}

\paragraph{Video-guided Machine Translation (VMT)}
In Figure~\ref{fig:vmt_supp}, we showcase the advantages of the VMT model over the base neural machine translation (NMT) model. Moreover, we further conduct the masked machine translation experiments and qualitatively demonstrate the effectiveness of VMT in recovering nouns or verbs in Figure~\ref{fig:mvmt_supp}.

\end{document}